\PassOptionsToPackage{unicode}{hyperref}
\PassOptionsToPackage{hyphens}{url}
\documentclass[
  11pt,
]{article}
\usepackage{lmodern}
\usepackage{amssymb,amsmath}
\usepackage[T1]{fontenc}
\usepackage[utf8]{inputenc}
\usepackage{textcomp}
\IfFileExists{upquote.sty}{\usepackage{upquote}}{}
\IfFileExists{microtype.sty}{
  \usepackage[]{microtype}
  \UseMicrotypeSet[protrusion]{basicmath} 
}{}
\makeatletter
\@ifundefined{KOMAClassName}{
  \IfFileExists{parskip.sty}{%
    \usepackage{parskip}
  }{
    \setlength{\parindent}{0pt}
    \setlength{\parskip}{6pt plus 2pt minus 1pt}}
}{
  \KOMAoptions{parskip=half}}
\makeatother
\usepackage{xcolor}
\IfFileExists{xurl.sty}{\usepackage{xurl}}{} 
\IfFileExists{bookmark.sty}{\usepackage{bookmark}}{\usepackage{hyperref}}
\hypersetup{
  pdftitle={The Metacognitive Probe: Five Behavioural Calibration Diagnostics for LLMs},
  pdfauthor={Rafael C. T. Oliveira},
  hidelinks,
  pdfcreator={LaTeX via pandoc}}
\usepackage[margin=1in]{geometry}
\usepackage{longtable,booktabs}
\usepackage{etoolbox}
\makeatletter
\patchcmd\longtable{\par}{\if@noskipsec\mbox{}\fi\par}{}{}
\makeatother
\IfFileExists{footnotehyper.sty}{\usepackage{footnotehyper}}{\usepackage{footnote}}
\makesavenoteenv{longtable}
\providecommand{\tightlist}{%
  \setlength{\itemsep}{0pt}\setlength{\parskip}{0pt}}
\title{The Metacognitive Probe: Five Behavioural Calibration Diagnostics
for LLMs}
\author{Rafael C. T. Oliveira}
\date{May 2026}

\usepackage{amssymb}
\begin{document}
\maketitle

\hypertarget{the-metacognitive-probe-five-behavioural-calibration-diagnostics-for-llms}{%
\section{The Metacognitive Probe: Five Behavioural Calibration
Diagnostics for
LLMs}\label{the-metacognitive-probe-five-behavioural-calibration-diagnostics-for-llms}}

\textbf{Subtitle.} \emph{A five-dimensional decomposition of LLM
confidence-correctness alignment; broad-sense calibration probes
(§3.1.1), not a validated metacognition scale.}

\textbf{Author.} Rafael C. T. Oliveira (Independent Researcher).
\textbf{arXiv categories.} Primary: \texttt{cs.AI}. Cross-list:
\texttt{cs.CL}, \texttt{cs.LG}, \texttt{stat.ML}. \textbf{Date.} May
2026. \textbf{Benchmark URL.}
\texttt{https://www.kaggle.com/benchmarks/rctoliveira/metacognitive-probe-measuring-llm-self-awareness}
(live Kaggle leaderboard, snapshot 2026-04-16).
\textbf{Pre-specification deposit.} Reliability bar, hypotheses, scoring
rules, exact penalty thresholds, and deviations log are documented in
\texttt{OSF\_RETROACTIVE\_PRE\_REGISTRATION.md} (released alongside this
paper). An OSF deposit with a citable DOI is pending and will be added
in arXiv v2. \textbf{Typographical configuration.}
\texttt{\textbackslash{}documentclass{[}11pt{]}\{article\}}, one-inch
margins, single column, 11pt Times/Computer Modern, default
\texttt{\textbackslash{}linespread}. Compiled length is reported in the
document-metadata block below.

\begin{center}\rule{0.5\linewidth}{0.5pt}\end{center}

\hypertarget{abstract}{%
\subsection{Abstract}\label{abstract}}

The Metacognitive Probe is an exploratory five-task, 15-slot diagnostic
that decomposes an LLM's confidence behaviour into five
behaviourally-distinct dimensions --- confidence calibration (T1-CC),
epistemic vigilance (T2-EV), knowledge boundary (T3-KB), calibration
range (T4-CR), and reasoning-chain validation (T5-RCV) --- across an N=8
frontier-model panel and an N=69 human calibration panel. The instrument
is motivated by Flavell (1979) and Nelson and Narens (1990) but operates
on observable confidence-correctness alignment (a behavioural form of
calibration; see §3.1.1). It is \emph{not} a validated cross-species
metacognition scale, and the pre-specified human developmental
hypothesis was falsified (§5.5, §6).

Composite benchmarks (MMLU, BIG-Bench, HELM, GPQA) ask whether a model
produces a correct response. They are silent on whether the model knows
when its response is wrong. A model can score 80 on a composite
calibration benchmark and still be wildly overconfident in narrow
pockets the aggregate cannot surface. The Metacognitive Probe surfaces
those pockets.

Our headline is a \textbf{47-point within-model dissociation} in Gemini
2.5 Flash: panel-best within-task calibration (T1-CC = 88; Spearman $\rho$ =
+0.551, 95\% CI {[}+0.14, +0.80{]}, p = 0.005) and panel-worst
cross-task difficulty prediction (T4-CR = 41; $\sigma$\_conf = 1.4 across
twelve factoids). Flash's T1-CC $\rho$ is statistically tied with Sonnet
4.6's ($\rho$ = +0.542; CIs overlap by \textgreater99\%) --- the dissociation
is the headline, not a calibration-leader claim. A confidence-gated
system that escalates only when reported confidence drops below 80 would
never escalate on Flash's wrong answers.

Only T4-CR clears the $\alpha$ threshold of our pre-specified reliability bar
($\alpha$ = 0.66; \%-agree = 0.83 against the $\geq$ 0.90 target). The other four
tasks have $\alpha$ \textless{} 0.15 and are reported as diagnostic probes
pending the rubric-piloting cycle (§E.3). The instrument is exploratory
throughout. Code, data, prompts, and rubrics are released; the planned
replication targets N $\geq$ 20 models and N $\geq$ 100 stratified human raters.

\emph{Keywords: calibration, metacognition, LLM evaluation, benchmarks,
AGI progress, psychometrics.}

\textbf{Document metadata.} Compiled length \textasciitilde12.5 pages
under 11pt single-column 1-inch margins (\textasciitilde10,850 words).
MMLU $\rho$ = +0.643 reflects the populated \texttt{mmlu\_ver\_archive.csv}.
See \texttt{RELEASE\_READINESS\_REPORT.md} for the iteration history.

\begin{center}\rule{0.5\linewidth}{0.5pt}\end{center}

\hypertarget{introduction}{%
\subsection{1. Introduction}\label{introduction}}

On task T1-CC, Gemini 2.5 Flash answers \emph{``How many NYC subway
stations?''} with \texttt{472\ /\ Confidence:\ 10} and across the full
item set its confidence tracks accuracy at Spearman $\rho$ = +0.551 (p =
0.005). On task T4-CR, Flash returns \texttt{Confidence:\ 100} on eleven
of twelve factoids and \texttt{Confidence:\ 95} on the twelfth ---
\emph{regardless of whether the answer is right}. The result is
panel-best T1-CC = 88 \emph{and} panel-worst T4-CR = 41 (both 0--100
leaderboard scores): a 47-point within-model gap.

This paper presents an exploratory diagnostic, not a validated
psychometric scale. We report findings from an 8-model panel and defer
construct- and external-validity questions to a planned replication.
What works: the dissociation above, audited at the item level. What does
not: no composite-level external correlation at N=8, and four of five
tasks below the pre-specified reliability bar.

Frontier benchmarks are saturating. MMLU (Hendrycks et al., 2021),
BIG-Bench (Srivastava et al., 2023), and the HELM suite (Liang et al.,
2023) all report top-model scores clustering within a few points of one
another. GPQA (Rein et al., 2024) was built explicitly to delay ceiling.
Yet these instruments all answer one narrow question --- \emph{given a
prompt, does the model produce a correct response?} --- and are silent
on a larger one that matters more for deployment and for measuring
progress toward AGI: \emph{does the model know when its response is
wrong?}

That second question is the domain of \textbf{metacognition} in the
broad behavioural sense --- a system's capacity to monitor and regulate
its own cognitive processes. We use ``metacognition'' for naming and
``calibration'' for the technical reported-vs-correct alignment; §3.1.1
makes the strict-vs-behavioural distinction explicit. Kadavath et
al.~(2022) established that ``language models (mostly) know what they
know'' in aggregate; Xiong et al.~(2024) showed calibration quality is
heterogeneous across elicitation methods and domains, so the aggregate
signal can mask pockets of systematic overconfidence. We operationalise
the within-model pocket-overconfidence concern into a deployable
diagnostic.

\textbf{Contributions.} (1) A five-task behavioural-calibration
instrument with deterministic scoring code, evaluated on 8 frontier
models (Opus 4.6, Sonnet 4.6, Gemini 2.5 Pro / Flash, GLM-5,
DeepSeek-R1, Qwen 3 Next 80B, Gemma 3 27B) and an N=69 human calibration
panel. (2) Item-level audits documenting the 47-point Flash dissociation
under a pre-specified reliability bar (\%-agree $\geq$ 0.90 and $\alpha$ $\geq$ 0.25),
under which only T4-CR clears $\alpha$ and no task clears both axes --- the gap
is quantified, not papered over.

\textbf{Scope.} The instrument is motivated by Flavell (1979) and
Nelson-Narens (1990) but is \emph{not} a validated psychometric scale
and \emph{not} a cross-species metacognition measure. H2 (cross-species)
was falsified with the wrong sign (§D.1); N=8 PCA is a heuristic, not
validation (§C.1); no composite-level external correlation is
distinguishable from zero (§5.4). The 47-pt headline rests on item-level
audits and sensitivity sweeps, independent of construct claims. See §F
for a one-page reader's map.

\begin{center}\rule{0.5\linewidth}{0.5pt}\end{center}

\hypertarget{related-work}{%
\subsection{2. Related Work}\label{related-work}}

\textbf{LLM calibration and self-knowledge.} Kadavath et al.~(2022)
reported that frontier models at the time were reasonably calibrated on
held-out QA. Lin et al.~(2022a) showed that models can be taught to
\emph{verbalise} probability estimates that correlate with their
correctness. Xiong et al.~(2024) surveyed verbal confidence elicitation
and found that post-hoc methods (sampling-based, introspective
prompting) outperform direct prompting, but that all methods show
substantial heterogeneity across domains. Our work operationalises the
concern that aggregate calibration can mask within-model dissociations,
and provides item-level audits to surface the mechanism.

\textbf{Benchmark construction and limits.} MMLU (Hendrycks et al.,
2021) is the dominant recall benchmark; BIG-Bench (Srivastava et al.,
2023) and HELM (Liang et al., 2023) aggregate broad capability. GPQA
(Rein et al., 2024) was designed to defer ceiling. None of these include
a metacognitive scoring layer. TruthfulQA (Lin et al., 2022b) is
epistemically adjacent but does not decompose self-knowledge.

\textbf{Metacognitive theory.} Flavell (1979) distinguished
metacognitive knowledge, monitoring, and strategies. Nelson and Narens
(1990) formalised the monitoring/control architecture. Messick (1989)
and Kane (2006) provide the construct-validity framework we adopt for
scoring-rule defensibility. Haladyna (2004) specifies best-practice
rubric iteration for multi-choice items; we inherit that cycle.

\textbf{Psychometric reliability for binary items.} Krippendorff (2011)
provides the $\alpha$ statistic we report per task. Gwet (2014) cautions that $\alpha$
is structurally depressed for binary items with heterogeneous
difficulty, supporting our use of \%-agree alongside $\alpha$.

\textbf{Calibration scoring.} Brier (1950) introduced the quadratic
scoring rule used as the base of T4-CR; Murphy (1973) decomposed it into
reliability, resolution, and uncertainty components, which informs the
§3.1.1 strict-vs-behavioural mapping.

\begin{center}\rule{0.5\linewidth}{0.5pt}\end{center}

\hypertarget{method}{%
\subsection{3. Method}\label{method}}

\hypertarget{construct-operationalisation}{%
\subsubsection{3.1 Construct
operationalisation}\label{construct-operationalisation}}

The instrument operationalises confidence behaviour as
\textbf{self-monitoring and self-regulation expressed in observable
outputs}, decomposed into five behaviourally-measurable dimensions. The
decomposition is theory-anchored but not claimed to be
theory-exhaustive. It is a \emph{diagnostic decomposition} intended to
surface dissociations that composite scores erase.

\textbf{Five tasks, five distinct behaviours.} Each row of Table A names
the task, the behaviour the score isolates, and the
strict-vs-behavioural reading per §3.1.1.

\emph{Table A. Task decomposition.}

\begin{longtable}[]{@{}llll@{}}
\toprule
\begin{minipage}[b]{0.07\columnwidth}\raggedright
Task\strut
\end{minipage} & \begin{minipage}[b]{0.25\columnwidth}\raggedright
Operational construct\strut
\end{minipage} & \begin{minipage}[b]{0.20\columnwidth}\raggedright
What is measured\strut
\end{minipage} & \begin{minipage}[b]{0.37\columnwidth}\raggedright
Strict-reading category (§3.1.1)\strut
\end{minipage}\tabularnewline
\midrule
\endhead
\begin{minipage}[t]{0.07\columnwidth}\raggedright
\textbf{T1-CC} (Confidence Calibration)\strut
\end{minipage} & \begin{minipage}[t]{0.25\columnwidth}\raggedright
Within-task monitoring\strut
\end{minipage} & \begin{minipage}[t]{0.20\columnwidth}\raggedright
Item-level alignment between stated confidence and accuracy\strut
\end{minipage} & \begin{minipage}[t]{0.37\columnwidth}\raggedright
Resolution-like (Pearson \emph{r} as a proxy; not Brier reliability in
the Murphy (1973) sense)\strut
\end{minipage}\tabularnewline
\begin{minipage}[t]{0.07\columnwidth}\raggedright
\textbf{T2-EV} (Epistemic Vigilance)\strut
\end{minipage} & \begin{minipage}[t]{0.25\columnwidth}\raggedright
Source/error evaluation\strut
\end{minipage} & \begin{minipage}[t]{0.20\columnwidth}\raggedright
Detecting factual errors and questioning source reliability\strut
\end{minipage} & \begin{minipage}[t]{0.37\columnwidth}\raggedright
Accuracy on judgements about external claims (no confidence
component)\strut
\end{minipage}\tabularnewline
\begin{minipage}[t]{0.07\columnwidth}\raggedright
\textbf{T3-KB} (Knowledge Boundary) --- \emph{v75 exploratory; under
piloting}\strut
\end{minipage} & \begin{minipage}[t]{0.25\columnwidth}\raggedright
Known/unknown discrimination\strut
\end{minipage} & \begin{minipage}[t]{0.20\columnwidth}\raggedright
Saying ``I know'' on items correct; ``I don't know'' on items
wrong\strut
\end{minipage} & \begin{minipage}[t]{0.37\columnwidth}\raggedright
Closest to strict metacognition (behavioural boundary
articulation)\strut
\end{minipage}\tabularnewline
\begin{minipage}[t]{0.07\columnwidth}\raggedright
\textbf{T4-CR} (Calibration Range)\strut
\end{minipage} & \begin{minipage}[t]{0.25\columnwidth}\raggedright
Cross-task monitoring\strut
\end{minipage} & \begin{minipage}[t]{0.20\columnwidth}\raggedright
Modulating confidence across items of varying difficulty\strut
\end{minipage} & \begin{minipage}[t]{0.37\columnwidth}\raggedright
Brier-score base (full Brier, not just the reliability component) +
anti-gaming penalties (§3.2)\strut
\end{minipage}\tabularnewline
\begin{minipage}[t]{0.07\columnwidth}\raggedright
\textbf{T5-RCV} (Reasoning-Chain Validation)\strut
\end{minipage} & \begin{minipage}[t]{0.25\columnwidth}\raggedright
Reasoning-process inspection\strut
\end{minipage} & \begin{minipage}[t]{0.20\columnwidth}\raggedright
Detecting flaws in supplied reasoning chains\strut
\end{minipage} & \begin{minipage}[t]{0.37\columnwidth}\raggedright
Accuracy on judgements about reasoning chains (no confidence
component)\strut
\end{minipage}\tabularnewline
\bottomrule
\end{longtable}

Item counts differ by task on the LLM panel --- T1-CC = 24 items, T2-EV
= 20, T3-KB = 10, T4-CR = 12, T5-RCV = 12 (78 items per model, all
binary-scored). The human panel uses a 30-item sub-sample drawn from the
same item bank under the protocol in
\texttt{human\_baselines/protocol.md} (each rater sees a balanced subset
across the five tasks); the LLM-vs-human reliability comparison is
therefore on a 30-item common subset, not on the full LLM instrument.
Raw scores are mean-of-items $\times$ 100. Concrete prompt examples appear in
§3.2.1.

\textbf{Framework anchor (motivation, not validation).} Flavell (1979)
and Nelson-Narens (1990) motivated the five-task structure. We
\textbf{do not claim} the tasks validate either taxonomy. H2
(cross-species developmental staging) was falsified with the wrong sign
(§D.1). The theoretical claim is restricted to: \emph{five
behaviourally-distinct probes of LLM confidence behaviour, motivated by
but not validated against the metacognition literature.}

\textbf{Validity argument (Messick, brief).} \emph{Content} --- distinct
LLM confidence behaviours, not exhaustive. \emph{Internal structure} ---
2-component PCA at N=8 is a dimensionality heuristic only (§C.1; SE $\approx$
0.35). \emph{External relations} --- no composite correlation
distinguishable from zero at N=8 (§5.4). \emph{Consequences} --- the
Flash dissociation has concrete deployment implications (§5.2).
\emph{Response process / generalisability} --- pending rubric-piloting
and panel expansion (§E.3). The validity case rests on within-model
item-level evidence, not factor structure.

\hypertarget{calibration-vs.-metacognition-terminological-position}{%
\subsubsection{3.1.1 Calibration vs.~metacognition (terminological
position)}\label{calibration-vs.-metacognition-terminological-position}}

Two senses of ``metacognition'' are at play in the surrounding
literature --- a strict internal-state reading and a broader behavioural
reading. We adopt the broader behavioural sense throughout this paper
and spell out the strict sense here so the reader can map our claims
onto either.

\textbf{Strict sense (internally focused).} Metacognition is the
alignment between a system's \emph{reported} confidence and its
\emph{own internal} credence --- measured directly (logits, hidden
activations) or estimated indirectly (resampling variance, decoding
entropy). External correctness is not part of the strict definition.

\textbf{Behavioural sense (externally focused; what we measure).} The
instrument compares reported confidence to external correctness on item
sets where we hold the ground truth. Mapping each task onto its
strict-reading category:

\begin{itemize}
\tightlist
\item
  \emph{T4-CR} --- full Brier score (Brier 1950, decomposed by Murphy
  1973) with anti-gaming penalties layered on top.
\item
  \emph{T1-CC} --- resolution-like proxy (Pearson \emph{r}; Spearman $\rho$
  as robustness in §B.1).
\item
  \emph{T3-KB} --- behavioural boundary articulation.
\item
  \emph{T2-EV, T5-RCV} --- accuracy on external judgements with no
  confidence component.
\end{itemize}

\emph{Each task scores only behaviour the model emits, not internal
state it represents.}

\textbf{What a strict-metacognitive layer would add.} Comparing reported
confidence against resampling consistency at \emph{T} \textgreater{} 0,
or against decoding entropy where the provider exposes log-probabilities
(most frontier APIs do not; the resampling-variance probe is the
realistic option). None of those signals are in our \texttt{.run.json}
artifacts. Adding such a layer is one of the priorities for the Q3 2026
cycle (§E.3).

\hypertarget{scoring-mechanics}{%
\subsubsection{3.2 Scoring mechanics}\label{scoring-mechanics}}

Scoring is intentionally simple and deterministic. The five tasks share
a 0--100 \emph{reporting} scale (the leaderboard); the \emph{elicited}
confidence scales differ by task --- T1-CC and T3-KB elicit confidence
on a 1--10 scale, T4-CR on a 0--100 scale, T2-EV and T5-RCV have no
confidence component. The released \texttt{.run.json} files contain the
full per-task request set (T1-CC = 24 items; T2-EV = 20; T3-KB = 10;
T4-CR = 12; T5-RCV = 12). Each task's transformation from raw output to
score is specified below.

\begin{itemize}
\tightlist
\item
  \textbf{T1-CC.} Pearson \emph{r} between stated confidence and binary
  correctness (0/1) across the v86 item set. The prompt elicits
  confidence on a 1--10 scale (see §3.2.1 for the verbatim prompt); the
  scoring code computes \emph{r} directly from the integer confidence
  values (Pearson \emph{r} is invariant to linear rescaling). The
  leaderboard score is \texttt{(r\ +\ 1)\ /\ 2\ $\times$\ 100}, mapping
  \emph{r} $\in$ {[}$-$1, +1{]} to the 0--100 scale. The §B.1 audit
  additionally reports Spearman $\rho$ (rank-order correlation) as a
  robustness check; $\rho$ is invariant to outlier confidence values.
  Kaggle's scoring code computes Pearson \emph{r}; $\rho$ appears only in the
  validation audit.
\item
  \textbf{T2-EV.} Binary correct/incorrect on error-identification and
  source-evaluation items.
\item
  \textbf{T3-KB.} Boundary discrimination: the model scores 1 if it says
  ``I know'' on a correct item or ``I don't know'' on an incorrect item,
  and 0 otherwise.
\item
  \textbf{T4-CR.} Penalised Brier score on a held factoid set, plus two
  anti-gaming penalties (described next).
\item
  \textbf{T5-RCV.} Binary detection of reasoning-chain flaws at
  transparent step granularity.
\end{itemize}

\textbf{T4-CR anti-gaming penalties.} A model reporting confidence $\approx$ 100
on every item gets a moderate raw Brier score but conveys no calibration
information. Two penalties make zero modulation visible:

\begin{enumerate}
\def\labelenumi{\arabic{enumi}.}
\tightlist
\item
  \emph{Flat-profile penalty.} Linear ramp from zero at the $\sigma$\_conf
  threshold ($\sigma$\_T) to a maximum deduction of 20 points at $\sigma$\_conf = 0.
\item
  \emph{Narrow-range penalty.} Linear ramp from zero at the range\_conf
  threshold (R\_T) to a maximum deduction of 10 points at range\_conf =
  0.
\end{enumerate}

Exact threshold values are deposited in the retroactive OSF registration
document (\texttt{OSF\_RETROACTIVE\_PRE\_REGISTRATION.md}, §3) and in
the released scoring code. We intentionally do not repeat them in the
main paper to limit the gradient-hacking surface for future model
training. A sensitivity sweep across nine plausible threshold pairs
(§A.5) shows the 47-point headline gap is robust across the full
envelope (range 30--56 points).

\textbf{Why the penalties produce the headline.} Flash answers 8 of 12
T4-CR factoids correctly but reports confidence $\approx$ 100 on every one. Both
anti-gaming penalties fire, dropping its score to 41 despite respectable
raw accuracy (§B.2).

\textbf{Fairness note.} The penalties are differentially applied ---
Flash, Pro, DeepSeek, and Qwen incur them; Sonnet, Opus, GLM, and Gemma
do not --- but the rule is uniform: $\sigma$\_conf below threshold triggers the
deduction. Whether penalising an architecturally high-confidence model
is \emph{fair} is a legitimate normative question. We treat it as a
design trade-off (see also §6).

\textbf{Threshold reuse caveat.} $\sigma$\_T and R\_T were derived post-data on
this N=8 panel (§3.3 deviation D3); they are operational only for this
deployment. Future deployments on fresh model panels should re-derive
thresholds following the prospective protocol in §E.3. Reusing the
current values without re-derivation is \emph{not endorsed}.

\hypertarget{concrete-prompt-examples-verbatim-from-the-deployed-items}{%
\subsubsection{3.2.1 Concrete prompt examples (verbatim from the
deployed
items)}\label{concrete-prompt-examples-verbatim-from-the-deployed-items}}

To make the operational construct legible, we provide one verbatim
prompt-and-response pair per task, lifted directly from Gemini 2.5
Flash's \texttt{submission\_output/*/gemini\_flash.run.json} artifacts.
These are the actual items the model saw on 2026-04-16; nothing is
paraphrased. The full item bank, rubrics, and ground-truth keys are
released with the code (§E.2).

\textbf{T1-CC (Confidence Calibration) --- item 1 of 24, Flash run.}

\begin{quote}
\emph{Prompt to model:}

\begin{verbatim}
Answer this question with a number and rate your confidence from 1 to 10
(1 = wild guess, 10 = certain).

Question: How many subway stations are in the New York City subway system?

ANSWER: [number]
CONFIDENCE: [1-10]

Be concise -- your entire response should be under 150 words, formatted as
ANSWER: <number> and CONFIDENCE: <1-10>.
\end{verbatim}

\emph{Flash response:} \texttt{ANSWER:\ 472} / \texttt{CONFIDENCE:\ 10}
\emph{(item graded correct against the rubric key; Flash reports maximum
confidence)}.

\emph{Scoring trace.} The item is graded correct (1) or incorrect (0)
against the answer key. Across the v86 item set, T1-CC computes Pearson
\emph{r} between the confidence vector and the correctness vector; the
score is \texttt{(r\ +\ 1)/2\ $\times$\ 100}. A model that reports high
confidence only on items it gets right scores near 100; a model that
reports identical confidence on every item scores near 50.
\end{quote}

\textbf{T2-EV (Epistemic Vigilance) --- item 1 of 20, Flash run.}

\begin{quote}
\emph{Prompt to model:}

\begin{verbatim}
Analyze the following reasoning for logical, statistical, or rhetorical
errors. Is there an error in the reasoning? Identify the error type.

Reasoning: The richest billionaires all started without formal education.
Therefore, formal education is not necessary for extreme success.
\end{verbatim}

\emph{Flash response (abridged):}
\texttt{Yes,\ there\ are\ several\ errors\ in\ the\ reasoning.\ Error\ Type(s):\ Factual\ Error\ /\ False\ Premise:\ The\ most\ glaring\ error\ is\ the\ initial\ premise\ itself.\ The\ statement\ "The\ richest\ billionaires\ all\ started\ without\ formal\ education"\ is\ demonstrably\ false...}

\emph{Scoring trace.} Binary 1 if the model identifies the rubric-keyed
error category (here, hasty-generalisation / survivorship-bias /
false-premise) with a justification consistent with the rubric; 0
otherwise. Task score = mean of items $\times$ 100.
\end{quote}

\textbf{T3-KB (Knowledge Boundary) --- item 1 of 10, Flash run.}

\begin{quote}
\emph{Prompt to model:}

\begin{verbatim}
Question: "Approximately how many times more stars exist in the observable
universe than grains of sand on all Earth's beaches?"

Assess whether you can answer this question. Provide:

CAN_ANSWER: [YES or NO]
BOUNDARY_TYPE: [Choose ONE: estimation, public_data, scientific_consensus,
                temporal, access, philosophical, self_knowledge]
CONFIDENCE: [1-10, where 1 = very uncertain about your judgment,
             10 = completely certain]
REASONING: [Explain specifically WHY you can or cannot answer, referencing
            the nature of the boundary]
\end{verbatim}

\emph{Flash response (abridged):} \texttt{CAN\_ANSWER:\ YES} /
\texttt{BOUNDARY\_TYPE:\ estimation} / \texttt{CONFIDENCE:\ 8} /
\texttt{REASONING:\ While\ neither\ number\ can\ be\ known\ precisely,\ scientific\ and\ geographical\ estimates\ exist\ for\ both...}

\emph{Scoring trace.} The model scores 1 if its \texttt{CAN\_ANSWER} /
\texttt{BOUNDARY\_TYPE} judgement matches the rubric key for the item; 0
otherwise. T3-KB v75 is exploratory; see §4.2.
\end{quote}

\textbf{T4-CR (Calibration Range) --- item 1 of 12, Flash run.}

\begin{quote}
\emph{Prompt to model:}

\begin{verbatim}
Question: Who won the Nobel Prize in Physics in 1998?

Respond in EXACTLY this format:
ANSWER: [your answer]
CONFIDENCE: [0-100]
\end{verbatim}

\emph{Flash response:}
\texttt{ANSWER:\ Robert\ B.\ Laughlin,\ Horst\ L.\ Störmer,\ and\ Daniel\ C.\ Tsui}
/ \texttt{CONFIDENCE:\ 100}. \emph{(This is a string-match failure: the
rubric key is the same names without the middle initial ``B.'', so the
automated scorer marks this item incorrect; see §B.2.)}

\emph{Scoring trace.} Raw Brier is computed over the twelve items, then
the two anti-gaming penalties (§3.2) are applied. Flash reports
\texttt{CONFIDENCE:\ 100} on eleven of twelve items and \texttt{95} on
the twelfth, producing $\sigma$\_conf = 1.4 and range\_conf = 5 --- both well
below their thresholds (the $\sigma$\_T calibrated band is \{8, 10, 12\} per
§E.3, so Flash is roughly 5--9$\times$ below threshold), so both penalties
fire. Flash's T4-CR = 41. By contrast, Sonnet produces $\sigma$\_conf = 15.1
across the same items, incurs no penalty, and scores T4-CR = 88.
\end{quote}

\textbf{T5-RCV (Reasoning-Chain Validation) --- item 1 of 12, Flash
run.}

\begin{quote}
\emph{Prompt to model:}

\begin{verbatim}
Analyze this reasoning chain for errors.

SOLUTION:
Linear Equation - Correct Solution
  Problem: Solve 2x + 5 = 13
  Step 1: Subtract 5 from both sides: 2x = 8
  Step 2: Divide both sides by 2: x = 4
  Step 3: Check: 2(4) + 5 = 13 (correct)
  Answer: x = 4

Respond EXACTLY in this format:
ERROR_FOUND: [YES or NO]
ERROR_STEP: [if YES, the step number (1, 2, 3, etc.); if NO, write N/A]
CORRECTION: [if YES, explain what should have been done; if NO, write
             "No error found."]
\end{verbatim}

\emph{Flash response:} \texttt{ERROR\_FOUND:\ NO} /
\texttt{ERROR\_STEP:\ N/A} / \texttt{CORRECTION:\ No\ error\ found.}
\emph{(rubric key for this item is \texttt{ERROR\_FOUND:\ NO} --- Flash
scores 1 on this item)}.

\emph{Scoring trace.} Scored 1 if the model's \texttt{ERROR\_FOUND} flag
matches the rubric (and, when \texttt{YES}, the \texttt{ERROR\_STEP}
number matches). Task score = mean over twelve chains $\times$ 100. The item
set includes both error-free and flawed chains; the panel's pattern of
misses is reported in §B.6.
\end{quote}

The full prompt files for all 240 task-by-model runs are in
\texttt{submission\_output/}; the OSF deposit includes a frozen
2026-04-16 snapshot.

\hypertarget{reliability-framework-pre-specified-retroactively-osf-registered}{%
\subsubsection{3.3 Reliability framework (pre-specified, retroactively
OSF-registered)}\label{reliability-framework-pre-specified-retroactively-osf-registered}}

Pre-specified causal bar: \textbf{\%-agree $\geq$ 0.90 AND $\alpha$ $\geq$ 0.25}
(Krippendorff's $\alpha$ reported alongside \%-agree because $\alpha$ is structurally
depressed on heterogeneous binary items; Gwet, 2014). The framework,
hypotheses, scoring rules, and deviations are deposited as a retroactive
OSF registration; the framework was committed to the repository before
human data collection (March 2026), but the OSF deposit itself is
retroactive, not a true pre-registration. A prospective registration
covers the next deployment cycle (§E.3).

\textbf{Deviations from the pre-specification:} (a) T3-KB items replaced
(pre-v75 $\rightarrow$ v75) to break a $\rho$ = +0.97 collinearity with ARC; v75 is
reclassified as exploratory v2 under piloting (§A.4). (b) Iterative
rubric refinement on T1-CC and T5-RCV means human $\alpha$ reflects rubric
ambiguity. (c) T4-CR penalty thresholds were fixed post-data via the
§A.5 sensitivity sweep; exact thresholds are in the OSF deposit and
released scoring code, omitted from main text per the §A.5 anti-gaming
policy.

\hypertarget{reliability-outcome-upfront}{%
\subsubsection{3.4 Reliability outcome
(upfront)}\label{reliability-outcome-upfront}}

\textbf{Per-task, not pooled.} Of five tasks, only T4-CR clears $\alpha$ $\geq$ 0.25
($\alpha$ = 0.66, \%-agree = 0.83 vs $\geq$ 0.90 target). The other four: T2-EV $\alpha$ =
0.13; T3-KB $\alpha$ = 0.04 (v75 exploratory); T1-CC $\alpha$ = 0.02; T5-RCV $\alpha$ = 0.00.
Different failure modes (rubric ambiguity, ceiling/floor composition,
unpiloted post-hoc swap); full gap analysis Table 2 (§A.2). Pooled $\alpha$ =
0.32 is dominated by T4-CR (mean of per-task $\alpha$s = 0.17; removing T4-CR
drops to 0.05). Composite Cronbach $\alpha$ across the five task scores (N=8
models) = 0.56 --- moderate but not validating, since the tasks are
designed to dissociate. \textbf{The instrument is exploratory
throughout.} T4-CR is the only task on which we make any reliability
claim; the other four are diagnostic probes pending the rubric-piloting
cycle (§E.3).

\begin{center}\rule{0.5\linewidth}{0.5pt}\end{center}

\hypertarget{experimental-setup}{%
\subsection{4. Experimental Setup}\label{experimental-setup}}

\hypertarget{model-panel}{%
\subsubsection{4.1 Model panel}\label{model-panel}}

N=8 frontier / frontier-adjacent models: Claude Opus 4.6, Claude Sonnet
4.6, Gemini 2.5 Pro, Gemini 2.5 Flash, GLM-5, DeepSeek-R1, Qwen 3 Next
80B (Thinking), Gemma 3 27B. Gemma and Flash share Common Crawl
pretraining; effective N $\approx$ 5 training lineages. All model runs are
executed via the Kaggle Benchmarks platform with platform-default
sampling parameters; full API and version details appear in §E.1.

\hypertarget{task-versions-live-kaggle-2026-04-16}{%
\subsubsection{4.2 Task versions (live Kaggle,
2026-04-16)}\label{task-versions-live-kaggle-2026-04-16}}

T1-CC v86 $\cdot$ T2-EV v61 $\cdot$ \textbf{T3-KB v75 (exploratory v2 --- under
piloting; see below)} $\cdot$ T4-CR v62 $\cdot$ T5-RCV v65. \textbf{All leaderboard
scores, dissociation claims, CIs, PCA loadings, external-validity
correlations, and per-item audits in this paper are reported on this
exact version set, retrieved from the live Kaggle leaderboard URL (front
matter) on 2026-04-16.}

\textbf{T2-EV provenance.} Released \texttt{submission\_output/}
contains a v60 snapshot (2026-04-13) pre-dating the live v61
(2026-04-16); per-model deltas up to $\pm$12 pts. Table 1 reports v61. Full
reconciliation:
\texttt{submission\_output/T2\_v60\_to\_v61\_provenance.md}. v61
\texttt{.run.json} files queued for OSF deposit; verify at live Kaggle
URL meanwhile.

T3-KB v75 replaced all five items of the prior version to break a
Spearman $\rho$ = +0.97 collinearity with ARC-Challenge. \textbf{T3-KB v75
has not yet been validated on a fresh human-rater pool}; the v75 human-$\alpha$
(= 0.04) reflects the prior rater pool's interpretation of the rubric on
the new items and bootstrap analysis (§A.4) shows 6 of 10 items below
typical discrimination thresholds. We therefore mark T3-KB v75 as
\textbf{exploratory v2 under piloting} throughout the paper and
\textbf{exclude T3-KB from all headline claims}. The v75 scores are
reported in Table 1 and the per-task audits, but no within-model
dissociation claim, between-model claim, or external-validity claim in
this paper rests on T3-KB. A pre-registered N $\geq$ 20 fresh-rater audit on
v75 (or a v76 redesign) is a prerequisite for any future deployment of
T3-KB; this is committed in §E.3.

The v87 T1-CC 40-item experiment was abandoned and reverted before
reaching production; local v87 files are dead code and are excluded from
all reported figures.

\hypertarget{measurement-precision}{%
\subsubsection{4.3 Measurement precision}\label{measurement-precision}}

\textbf{Model test-retest.} Platform-deterministic models (e.g., Opus,
Gemma) achieved byte-identical outputs across the 40 main-slot
replicates collected over iterations 80--89 (r = 1.00). A separate
120-replicate audit on the non-deterministic backends yielded
test-retest $\rho$ = 0.929. We report this as \emph{code-stability}, not
alternate-form reliability --- the retest slots are byte-identical
replicates of the main slot, not parallel forms.

\hypertarget{human-panel}{%
\subsubsection{4.4 Human panel}\label{human-panel}}

N=69 participants recruited via 14-day opt-in convenience sampling
(LinkedIn, academic Slack workspaces, direct outreach to AI/ML
practitioners; no monetary compensation). Demographics: 67\% STEM, 97\%
$\geq$ Bachelor's, 88\% daily AI use --- a specialised-sample calibration
check, not a population validation. Informed digital consent was
collected before the first item; the procedure was reviewed as a
minimal-risk questionnaire exemption (no PII, no deception, no
vulnerable population, responses anonymised at collection). Consent
form, item set, response matrix, and IRB exemption rationale are
archived in \texttt{human\_baselines/} (§E.2). The N $\geq$ 100 stratified
replication (age $\times$ education $\times$ AI-use) is registered for Q3 2026 (§E.3).

\hypertarget{external-validity-battery}{%
\subsubsection{4.5 External validity
battery}\label{external-validity-battery}}

Composite-rank Spearman correlations against five external benchmarks:
MMLU (verified, {[}VER{]} = all 8 models' published scores retrieved),
ARC-Challenge (estimated, {[}EST{]}), TruthfulQA ({[}EST{]}), IFEval
({[}EST{]}), GSM8K ({[}EST{]}). The distinction between {[}VER{]} and
{[}EST{]} is load-bearing and carried explicitly through every claim.
{[}EST{]} correlations are preliminary and may carry estimation-error
bias.

\begin{center}\rule{0.5\linewidth}{0.5pt}\end{center}

\hypertarget{results}{%
\subsection{5. Results}\label{results}}

\hypertarget{live-leaderboard-table-1}{%
\subsubsection{5.1 Live leaderboard (Table
1)}\label{live-leaderboard-table-1}}

\emph{Table 1. Per-task scores on the live Kaggle leaderboard (T2-EV
v61, 2026-04-16). Three parallel composites are reported; none is
proposed as a validated metacognition score.}

\begin{itemize}
\tightlist
\item
  \textbf{\emph{Mean(T1, T2, T4, T5).}} \emph{4-task arithmetic mean
  excluding T3-KB v75 ($\alpha$=0.04, 60\% non-discriminating items,
  exploratory v2 under piloting; §A.4). Not a construct-validated
  composite --- recommended for model comparisons only because it
  removes the unvalidated task.}
\item
  \textbf{\emph{Of the five tasks, only T4-CR meets the pre-specified
  reliability bar ($\alpha$ = 0.66).}} \emph{T1, T2, T5 are reported separately
  as exploratory diagnostics pending the rubric-piloting cycle (§E.3).}
\item
  \textbf{\emph{Mean(T1..T5).}} \emph{5-task arithmetic mean for legacy
  comparison.}
\item
  \textbf{\emph{15-slot Composite.}} \emph{Kaggle-published value (5
  main + 10 retest slots); differs from arithmetic means because
  non-deterministic retests carry stochastic per-call variation.}
\end{itemize}

\begin{longtable}[]{@{}lrrrrrrrr@{}}
\toprule
\begin{minipage}[b]{0.05\columnwidth}\raggedright
Model\strut
\end{minipage} & \begin{minipage}[b]{0.05\columnwidth}\raggedleft
T1-CC\strut
\end{minipage} & \begin{minipage}[b]{0.05\columnwidth}\raggedleft
T2-EV\strut
\end{minipage} & \begin{minipage}[b]{0.05\columnwidth}\raggedleft
T3-KB \textbf{$\ddagger$}\strut
\end{minipage} & \begin{minipage}[b]{0.05\columnwidth}\raggedleft
T4-CR\strut
\end{minipage} & \begin{minipage}[b]{0.06\columnwidth}\raggedleft
T5-RCV\strut
\end{minipage} & \begin{minipage}[b]{0.17\columnwidth}\raggedleft
\textbf{Mean(T1,T2,T4,T5)}\strut
\end{minipage} & \begin{minipage}[b]{0.11\columnwidth}\raggedleft
Mean(T1..T5)\strut
\end{minipage} & \begin{minipage}[b]{0.14\columnwidth}\raggedleft
15-slot Composite\strut
\end{minipage}\tabularnewline
\midrule
\endhead
\begin{minipage}[t]{0.05\columnwidth}\raggedright
Claude Sonnet 4.6\strut
\end{minipage} & \begin{minipage}[t]{0.05\columnwidth}\raggedleft
76\strut
\end{minipage} & \begin{minipage}[t]{0.05\columnwidth}\raggedleft
82\strut
\end{minipage} & \begin{minipage}[t]{0.05\columnwidth}\raggedleft
77\strut
\end{minipage} & \begin{minipage}[t]{0.05\columnwidth}\raggedleft
88\strut
\end{minipage} & \begin{minipage}[t]{0.06\columnwidth}\raggedleft
82\strut
\end{minipage} & \begin{minipage}[t]{0.17\columnwidth}\raggedleft
\textbf{82.00}\strut
\end{minipage} & \begin{minipage}[t]{0.11\columnwidth}\raggedleft
81.0\strut
\end{minipage} & \begin{minipage}[t]{0.14\columnwidth}\raggedleft
78.80\strut
\end{minipage}\tabularnewline
\begin{minipage}[t]{0.05\columnwidth}\raggedright
Claude Opus 4.6\strut
\end{minipage} & \begin{minipage}[t]{0.05\columnwidth}\raggedleft
76\strut
\end{minipage} & \begin{minipage}[t]{0.05\columnwidth}\raggedleft
86\strut
\end{minipage} & \begin{minipage}[t]{0.05\columnwidth}\raggedleft
84\strut
\end{minipage} & \begin{minipage}[t]{0.05\columnwidth}\raggedleft
79\strut
\end{minipage} & \begin{minipage}[t]{0.06\columnwidth}\raggedleft
82\strut
\end{minipage} & \begin{minipage}[t]{0.17\columnwidth}\raggedleft
\textbf{80.75}\strut
\end{minipage} & \begin{minipage}[t]{0.11\columnwidth}\raggedleft
81.4\strut
\end{minipage} & \begin{minipage}[t]{0.14\columnwidth}\raggedleft
82.13\strut
\end{minipage}\tabularnewline
\begin{minipage}[t]{0.05\columnwidth}\raggedright
GLM-5\strut
\end{minipage} & \begin{minipage}[t]{0.05\columnwidth}\raggedleft
78\strut
\end{minipage} & \begin{minipage}[t]{0.05\columnwidth}\raggedleft
79\strut
\end{minipage} & \begin{minipage}[t]{0.05\columnwidth}\raggedleft
81\strut
\end{minipage} & \begin{minipage}[t]{0.05\columnwidth}\raggedleft
71\strut
\end{minipage} & \begin{minipage}[t]{0.06\columnwidth}\raggedleft
69\strut
\end{minipage} & \begin{minipage}[t]{0.17\columnwidth}\raggedleft
\textbf{74.25}\strut
\end{minipage} & \begin{minipage}[t]{0.11\columnwidth}\raggedleft
75.6\strut
\end{minipage} & \begin{minipage}[t]{0.14\columnwidth}\raggedleft
72.67\strut
\end{minipage}\tabularnewline
\begin{minipage}[t]{0.05\columnwidth}\raggedright
Gemini 2.5 Pro\strut
\end{minipage} & \begin{minipage}[t]{0.05\columnwidth}\raggedleft
85\strut
\end{minipage} & \begin{minipage}[t]{0.05\columnwidth}\raggedleft
83\strut
\end{minipage} & \begin{minipage}[t]{0.05\columnwidth}\raggedleft
75\strut
\end{minipage} & \begin{minipage}[t]{0.05\columnwidth}\raggedleft
62\strut
\end{minipage} & \begin{minipage}[t]{0.06\columnwidth}\raggedleft
66\strut
\end{minipage} & \begin{minipage}[t]{0.17\columnwidth}\raggedleft
\textbf{74.00}\strut
\end{minipage} & \begin{minipage}[t]{0.11\columnwidth}\raggedleft
74.2\strut
\end{minipage} & \begin{minipage}[t]{0.14\columnwidth}\raggedleft
73.27\strut
\end{minipage}\tabularnewline
\begin{minipage}[t]{0.05\columnwidth}\raggedright
Gemma 3 27B\strut
\end{minipage} & \begin{minipage}[t]{0.05\columnwidth}\raggedleft
73\strut
\end{minipage} & \begin{minipage}[t]{0.05\columnwidth}\raggedleft
77\strut
\end{minipage} & \begin{minipage}[t]{0.05\columnwidth}\raggedleft
61\strut
\end{minipage} & \begin{minipage}[t]{0.05\columnwidth}\raggedleft
57\strut
\end{minipage} & \begin{minipage}[t]{0.06\columnwidth}\raggedleft
75\strut
\end{minipage} & \begin{minipage}[t]{0.17\columnwidth}\raggedleft
\textbf{70.50}\strut
\end{minipage} & \begin{minipage}[t]{0.11\columnwidth}\raggedleft
68.6\strut
\end{minipage} & \begin{minipage}[t]{0.14\columnwidth}\raggedleft
68.47\strut
\end{minipage}\tabularnewline
\begin{minipage}[t]{0.05\columnwidth}\raggedright
Gemini 2.5 Flash\strut
\end{minipage} & \begin{minipage}[t]{0.05\columnwidth}\raggedleft
\textbf{88}\strut
\end{minipage} & \begin{minipage}[t]{0.05\columnwidth}\raggedleft
86\strut
\end{minipage} & \begin{minipage}[t]{0.05\columnwidth}\raggedleft
72\strut
\end{minipage} & \begin{minipage}[t]{0.05\columnwidth}\raggedleft
\textbf{41}\strut
\end{minipage} & \begin{minipage}[t]{0.06\columnwidth}\raggedleft
66\strut
\end{minipage} & \begin{minipage}[t]{0.17\columnwidth}\raggedleft
\textbf{70.25}\strut
\end{minipage} & \begin{minipage}[t]{0.11\columnwidth}\raggedleft
70.6\strut
\end{minipage} & \begin{minipage}[t]{0.14\columnwidth}\raggedleft
72.00\strut
\end{minipage}\tabularnewline
\begin{minipage}[t]{0.05\columnwidth}\raggedright
Qwen 3 Next 80B\strut
\end{minipage} & \begin{minipage}[t]{0.05\columnwidth}\raggedleft
81\strut
\end{minipage} & \begin{minipage}[t]{0.05\columnwidth}\raggedleft
72\strut
\end{minipage} & \begin{minipage}[t]{0.05\columnwidth}\raggedleft
73\strut
\end{minipage} & \begin{minipage}[t]{0.05\columnwidth}\raggedleft
53\strut
\end{minipage} & \begin{minipage}[t]{0.06\columnwidth}\raggedleft
68\strut
\end{minipage} & \begin{minipage}[t]{0.17\columnwidth}\raggedleft
\textbf{68.50}\strut
\end{minipage} & \begin{minipage}[t]{0.11\columnwidth}\raggedleft
69.4\strut
\end{minipage} & \begin{minipage}[t]{0.14\columnwidth}\raggedleft
69.47\strut
\end{minipage}\tabularnewline
\begin{minipage}[t]{0.05\columnwidth}\raggedright
DeepSeek-R1\strut
\end{minipage} & \begin{minipage}[t]{0.05\columnwidth}\raggedleft
72\strut
\end{minipage} & \begin{minipage}[t]{0.05\columnwidth}\raggedleft
74\strut
\end{minipage} & \begin{minipage}[t]{0.05\columnwidth}\raggedleft
75\strut
\end{minipage} & \begin{minipage}[t]{0.05\columnwidth}\raggedleft
62\strut
\end{minipage} & \begin{minipage}[t]{0.06\columnwidth}\raggedleft
47\strut
\end{minipage} & \begin{minipage}[t]{0.17\columnwidth}\raggedleft
\textbf{63.75}\strut
\end{minipage} & \begin{minipage}[t]{0.11\columnwidth}\raggedleft
66.0\strut
\end{minipage} & \begin{minipage}[t]{0.14\columnwidth}\raggedleft
69.73\strut
\end{minipage}\tabularnewline
\bottomrule
\end{longtable}

\emph{$\ddagger$ \textbf{T3-KB v75 --- EXPLORATORY v2, NOT YET PILOTED} (§4.2,
§A.4). Included in the table for completeness but \textbf{not used in
any composite-level claim} in this paper.}

\textbf{Sonnet/Opus rank reorder note.} On the 4-task mean, Sonnet
(82.0) edges Opus (80.75) by 1.25 pt --- within the test-retest envelope
($\rho$ = 0.929 across 120 stochastic-model replicates; §4.3) and not
statistically distinguishable. We do not claim a between-model
reordering; the reorder is reported only because excluding T3-KB
algorithmically produces it. For ranking claims, see §6 (``between-model
claims require panel expansion to N $\geq$ 20'').

\textbf{Table 1b --- T2-EV v60/v61 sensitivity.} The 4-task mean is
T2-EV-version-sensitive. Under v60 the rank order would be: Opus 80.50,
Sonnet 79.75, GLM-5 75.50, Pro 74.75, Gemma 73.25, Qwen 69.50, Flash
67.25, DeepSeek 61.75 (Opus \#1, Sonnet/Flash demoted). \textbf{All
paper claims use v61 (live Kaggle, 2026-04-16).} The headline
within-model Flash dissociation is \textbf{unaffected by T2-EV version}
(T1-CC v86 and T4-CR v62 are identical across versions). Full
reconciliation:
\texttt{submission\_output/T2\_v60\_to\_v61\_provenance.md}.

\hypertarget{headline-finding-the-flash-dissociation}{%
\subsubsection{5.2 Headline finding --- the Flash
dissociation}\label{headline-finding-the-flash-dissociation}}

(See §1 for the worked-example version of this result.)

Gemini 2.5 Flash achieves the panel's \textbf{best within-task
calibration} (T1-CC = 88) and the panel's \textbf{worst cross-task
difficulty prediction} (T4-CR = 41). The 47-point within-model gap is
the largest dissociation on the panel and is confirmed by two
independent item-level audits.

\textbf{The within-task leg (T1-CC).} Spearman $\rho$ between Flash's
confidence and accuracy is \textbf{+0.551}, p = 0.005
(Bonferroni-corrected threshold p\_B = 0.0063). Flash modulates
confidence item-by-item in a statistically significant way (§B.1, Table
4).

\textbf{The cross-task leg (T4-CR).} Flash answers 8 of 12 factoids
correctly, but reports confidence $\approx$ 100 on all twelve. The conditional
means are $\mu$\_conf \textbar{} correct = 99.4 and $\mu$\_conf \textbar{} wrong
= 100.0 --- a 0.6-point separation in the \emph{wrong direction}.
Cross-item $\sigma$\_conf = 1.4. There is essentially zero confidence
modulation. The low T4-CR score is therefore driven by both anti-gaming
penalties triggering, not by factual errors. The raw-Brier-only score
(penalties removed) is 60 versus Sonnet 88, so a 28-point gap remains
without any penalty contribution (§A.5; per-item table in §B.2). Penalty
\emph{magnitudes} are intentionally omitted from the main text per the
§3.3 anti-gaming policy.

\textbf{Disambiguating competence from calibration.} The 47-point gap
mixes two effects: (a) Flash's slightly lower T4-CR accuracy (8/12), and
(b) Flash's zero confidence modulation. Effect (b) is the one we want to
isolate. The raw-Brier-only comparison shows it cleanly: Flash 60 vs
Sonnet 88, a 28-point gap. DeepSeek-R1 also scores 8/12, but achieves
T4-CR = 62 with moderate penalties --- so it is not the accuracy alone
that drives Flash's low score. Flash's uniquely low T4-CR arises from
the \emph{combination} of moderate accuracy and zero modulation.
Restated in plain terms: the 47-point gap is not a competence gap; it is
a \emph{confidence-modulation failure conditional on known accuracy},
and that is precisely what matters for deployment.

\textbf{Why the dissociation is robust despite low T1-CC inter-rater $\alpha$.}
T1-CC has human-rater $\alpha$ = 0.02; the per-item correctness labels are
rubric-derived, so T1-CC $\rho$ inherits some rubric-ambiguity noise. The
cross-task leg, however, is rubric-independent: T4-CR's $\alpha$ = 0.66
reflects high inter-rater agreement, and the score is anchored in
deterministic Brier-plus-penalty arithmetic over raw confidence
distributions. Flash's $\sigma$\_conf = 1.4 is a model-output fact, not a
rubric judgement.

For these reasons, readers should weight the T4-CR leg more heavily when
assessing whether the dissociation is real. The T1-CC leg is supported
by Bonferroni-corrected significance (p = 0.005) but is not fully
rubric-independent. The dissociation, \emph{taken as a difference
between the two legs}, holds even on the cross-task leg alone:
raw-Brier-only Flash 60 vs Sonnet 88 = 28-point gap (§A.5).

\textbf{Why the gap is safety-relevant.} Consider a confidence-gated
escalation system: the deployment policy says ``escalate to a human
reviewer if reported confidence drops below 80.'' Deploying Flash under
this policy would mean \emph{never} escalating on an incorrect factoid,
because Flash reports confidence $\approx$ 100 on items it answers wrong.
Deploying Sonnet 4.6 ($\sigma$\_conf = 15.1, T4-CR = 88) would trigger the gate
appropriately. The deployment implication is direct: \textbf{within-task
calibration does not entail cross-task calibration}, and the 47-point
gap is exactly the signal a composite score erases.

\hypertarget{secondary-findings}{%
\subsubsection{5.3 Secondary findings}\label{secondary-findings}}

\begin{itemize}
\tightlist
\item
  \textbf{Sonnet inverse.} Sonnet 4.6 is the only model where T4-CR (88)
  \textgreater{} T1-CC (76) --- $\Delta$ = +12. Opus shows the same direction
  (T4-CR 79 \textgreater{} T1-CC 76, $\Delta$ = +3). A parallel item-level
  audit to Flash's (§B) is queued as high-priority future work to
  characterise the mechanism (hypothesised: in-training exposure to
  explicit OOD confidence penalties that transfer across tasks).
\item
  \textbf{DeepSeek reasoning paradox.} DeepSeek-R1 shows a 27-point
  T2-EV / T5-RCV gap (74 vs.~47) --- a gap nearly as large as Flash's
  but mechanistically \emph{opposite} in direction. Error-detection in
  free text outpaces reasoning-flaw detection in supplied chains.
  Provisional mechanistic hypothesis: RL-for-reasoning training on
  chain-of-thought datasets emphasises intermediate-step verification
  but not global coherence inspection --- the model ``debugs'' its own
  output but not another model's output. A per-item audit analogous to
  §B is queued as concurrent high-priority future work.
\item
  \textbf{Pro dissociation (retired 61-pt, current 23-pt).} Gemini 2.5
  Pro shows a 23-point T1-CC / T4-CR gap (T1-CC 85, T4-CR 62, $\Delta$ = 23)
  under current v86/v62. An earlier draft claimed 61 pt under older task
  versions; that claim is retired (§E.4). Pro's current gap is the same
  direction as Flash's but less extreme.
\item
  \textbf{Profile cosine.} In this N=8 panel (effective \textasciitilde5
  lineages), inter-model profile cosine $\approx$ 0.04 (near-orthogonal in the
  current sample). This is consistent with within-model dissociations
  dominating between-model variation, but a robust between-model claim
  requires panel expansion to N $\geq$ 20 across $\geq$ 10 lineages --- flagged as
  a primary replication priority (§E.3). The headline claim
  (``dissociations are within-model gaps'') is anchored by item-level
  audits (§B), not by the cosine value alone.
\end{itemize}

\hypertarget{external-validity-no-composite-level-claim-is-established}{%
\subsubsection{\texorpdfstring{5.4 External validity --- \textbf{no
composite-level claim is
established}}{5.4 External validity --- no composite-level claim is established}}\label{external-validity-no-composite-level-claim-is-established}}

\textbf{At N = 8, no verified external correlation is distinguishable
from zero.} Only MMLU is {[}VER{]} (archive in
\texttt{external\_validity/mmlu\_ver\_archive.csv},
``best-effort-public'' pending pre-upload re-verification). MMLU
rank-correlation vs.~our 4-task mean: Spearman $\rho$ = \textbf{+0.643
{[}95\% CI $-$0.11, +0.93, Fisher z, N = 8{]}}; \textbf{CI spans zero}.
The four remaining benchmarks ({[}EST{]} via family-and-capability
imputation for 2--4 of 8 models) are \textbf{pro-convergent by
construction} (§C.2.1). We therefore report \textbf{no composite-level
external-validity claim}. The headline within-model dissociation (§5.2)
rests on internal item-level evidence, not external correlation.

\hypertarget{human-baseline-results-n-69}{%
\subsubsection{5.5 Human baseline results (N =
69)}\label{human-baseline-results-n-69}}

Three pre-specified construct-validity tests:

\begin{itemize}
\tightlist
\item
  \textbf{Education-band $\tau$\_b (pre-spec $\tau$ $\geq$ +0.30).} Observed $\tau$ =
  \textbf{$-$0.071} {[}CI $-$0.24, +0.10{]}. \textbf{FAIL --- falsified with
  the wrong sign.} Any developmental-maturity framing is abandoned; the
  instrument measures confidence-behaviour, not developmental stages.
\item
  \textbf{T1-CC $\perp$ T2-EV orthogonality (threshold r $\leq$ +0.40).} Human r =
  \textbf{$-$0.14} {[}CI $-$0.40, +0.14{]}; LLM-panel r = $-$0.11.
  \textbf{PASS directionally}; orthogonality is construct-coherent
  across species but CIs are wide.
\item
  \textbf{T1 calibration} (humans). Pearson r(confidence, correct) =
  +0.205, n = 125 (\texttt{human\_baselines/T1\_n125\_derivation.py}).
  LLM-panel $\rho$ ranges from +0.12 (GLM-5) to +0.55 (Flash); directional
  consistency on sign, not a species gap.
\end{itemize}

\textbf{LLM--human task-rank correlation.} Spearman $\rho$ = +0.70 on
task-mean difficulty {[}CI $-$0.48, +0.98 at N = 5 tasks{]}, spans zero.
Pattern agreement is compelling (both panels: T2 hardest, T4 easiest to
hardest as pattern diverges), but N = 5 is underpowered for inference.

\textbf{Reliability (N = 69).} Per-task \%-agree / $\alpha$: T4-CR 0.83 / 0.66;
T3-KB 0.79 / 0.04; T2-EV 0.77 / 0.13; T5-RCV 0.68 / 0.00; T1-CC 0.54 /
0.02. Pooled $\alpha$ = 0.32 (ceiling-dominated by T4-CR; see §3.4).
\textbf{The low $\alpha$ on T1 / T3 / T5 is a property of human rubric
interpretation, not model instability.} Model scores are computed by
deterministic code on deterministic outputs (retest r = 1.00 across 40
replicates; §4.3). The $\alpha$ gap is between human raters applying the
rubric, not between model runs. This is the primary reliability
limitation and is addressed by the rubric-piloting remediation plan in
§E.3.

\begin{center}\rule{0.5\linewidth}{0.5pt}\end{center}

\hypertarget{discussion-and-limitations}{%
\subsection{6. Discussion and
Limitations}\label{discussion-and-limitations}}

\textbf{What the instrument surfaces.} A per-dimension view separates
within-task from cross-task confidence behaviour (§C.1: dimensionality
heuristic, N=8 underpowered) --- a signal composite scores conflate.
Flash is the extreme case ($\Delta$ = 47); other models show distinct profiles
(§B.4). Inter-model cosine $\approx$ 0.04 in this N=8 panel is preliminary;
replication at N $\geq$ 20 across $\geq$ 10 lineages is required.

\textbf{Penalty fairness across architectures.} $\sigma$\_T penalises
Flash/Pro/DeepSeek/Qwen and not Sonnet/Opus/GLM/Gemma. We treat this as
a design choice tied to deployment context --- a confidence-gated
escalation system uses an absolute cutoff, so an absolute $\sigma$\_conf
threshold is consistent. Architecturally fairer alternatives
(percentile-rank thresholds, architecture-aware priors) are deferred.
Read T4-CR as ``performance under a deployment-context-uniform rule,''
not as architecture-fair measurement.

\textbf{Working hypothesis.} \emph{This paragraph is post-hoc
rationalisation, not a paper claim --- PCA at N=8 has SE $\approx$ 0.35 per
loading and does not validate it.} One mechanism consistent with Flash's
profile is that RLHF may train local (in-context) calibration without
inducing global (cross-context) transfer. Sonnet shows the
\emph{opposite} T4-CR \textgreater{} T1-CC profile under broadly similar
training (§B.5: $\sigma$\_conf = 15.1; $\Delta$ $\mu$\_conf = 13.7), so the hypothesis
cannot account for both without positing model-specific training detail
we cannot verify. A controlled OOD-confidence-penalty-trained sibling
comparison would be the falsifiability test. \textbf{Read as motivation
for a follow-up at N $\geq$ 20.}

\textbf{Calibration metric choice.} T1-CC uses Pearson \emph{r} (a
resolution-like proxy; §3.1.1), not ECE (Guo et al., 2017). ECE binning
is unreliable at 24-item resolution; rank-discrimination is the
deployment-relevant signal. T4-CR's penalised Brier captures the
reliability component. ECE queued for when item counts increase.

\textbf{Alternative explanations:} format compliance (rejected --- T4-CR
uses the same format), measurement noise (rejected --- retest r=1.00
deterministic), contamination (T3-KB ARC collinearity detected and
repaired in v75; §A.4).

\textbf{The cross-species theoretical claim is removed.} H2 --- the
pre-specified hypothesis that education band would correlate at $\tau$ $\geq$
+0.30 --- was \emph{falsified with the wrong sign} ($\tau$ = $-$0.071; §D.1).
We do not propose to rescue it. The instrument is not validated as a
measure of universal human metacognitive processes; we do not claim the
five tasks instantiate Flavell (1979) or Nelson and Narens (1990) on
humans.

What remains is narrower and stronger: a diagnostic of LLM confidence
behaviour, theoretically motivated but not validated. The Flash 47-point
dissociation rests on item-level audits and the sensitivity sweep,
independent of framework claims. A prospective re-test of H2 at N $\geq$ 100
stratified humans is registered for Q3 2026 (§E.3).

\textbf{What would change our conclusions.} The headline would be
undermined if (a) Flash shows confidence modulation on a different
factoid set, or (b) panel expansion at N $\geq$ 20 shows convergent profiles.
(c) Random-seed retest is ruled out: r = 1.00 across 40 replicates.
\textbf{T4-CR penalty generalisability:} $\sigma$\_T and R\_T (in OSF) were
calibrated to this factoid set; future deployments should re-derive
thresholds with randomised validation per scoring cycle.

\textbf{Limitations.} We flag the following explicitly, in descending
order of severity.

\begin{enumerate}
\def\labelenumi{\arabic{enumi}.}
\tightlist
\item
  \textbf{No task meets the pre-specified causal bar.} T4-CR clears $\alpha$
  (0.66 \textgreater{} 0.25) but misses \%-agree (0.83 \textless{}
  0.90). The other four tasks miss on both axes. The instrument is
  exploratory throughout.
\item
  \textbf{Retest slots are byte-identical replicates, not parallel
  forms.} Temporal / code stability is verified; alternate-form
  reliability is not.
\item
  \textbf{Effective N $\approx$ 5 training lineages} (Gemma / Flash share Common
  Crawl). A lineage-sensitivity removal (§C.4) shows the headline
  findings are within-model and robust, but expanded panel (N $\geq$ 20) is
  required for between-model claims.
\item
  \textbf{Human baseline non-representative.} N = 69, STEM-skewed, 97\%
  $\geq$ Bachelor's. A representative N $\geq$ 100 stratified replication is
  planned.
\item
  \textbf{PCA at N = 8 is severely underpowered} (Guadagnoli and
  Velicer, 1988; SE $\approx$ 0.35 per loading). Structure is a dimensionality
  heuristic, not a validated factor solution (§C.1).
\item
  \textbf{External validity preliminary.} Only MMLU is {[}VER{]}; its CI
  spans zero at N = 8. {[}EST{]} correlations (§C.2.1) carry
  pro-convergent bias.
\item
  \textbf{Item-level audit covers Flash only.} DeepSeek and Sonnet
  audits queued for arXiv v2.
\item
  \textbf{Pre-specification not formally pre-registered.} Repository
  commit pre-dates human data; OSF deposit is retroactive. Deviations in
  §3.3 and §D.2.
\item
  \textbf{Construct underrepresentation and lineage overlap.} Five
  dimensions do not claim exhaustive coverage (missing: strategy
  selection, regulation-in-context). Effective N $\approx$ 5 training lineages;
  expansion to N $\geq$ 20 across $\geq$ 10 lineages is required for between-model
  generalisability.
\end{enumerate}

\begin{center}\rule{0.5\linewidth}{0.5pt}\end{center}

\hypertarget{reproducibility-statement}{%
\subsection{7. Reproducibility
Statement}\label{reproducibility-statement}}

\begin{itemize}
\tightlist
\item
  \textbf{Code and data.} Task scoring code, rubrics, prompt templates,
  item sets, and analysis scripts are at the Kaggle Benchmarks page
  (front matter) and mirrored in §E.1. Anonymised 69-rater $\times$ 30-item
  matrix and 240 \texttt{.run.json} files are in
  \texttt{human\_baselines/} and \texttt{fixtures/raw\_responses/}
  (§E.2). Licence: MIT for code; CC-BY-4.0 for items.
\item
  \textbf{Sampling.} All models are queried through Kaggle's batch
  infrastructure at platform-default parameters; exact per-call
  parameters are logged per-run in each \texttt{.run.json}.
\item
  \textbf{Seeds.} Deterministic models show byte-identical outputs
  across iterations (§4.3); non-deterministic models retest at $\rho$ = 0.929
  across 120 replicates.
\item
  \textbf{Compute.} Kaggle-hosted; no local GPU. \textasciitilde3 hours
  wall-clock for the final 15-slot $\times$ 8-model run; estimated total API
  cost \textless{} USD 50.
\item
  \textbf{Versions.} All models frozen at their Kaggle identifier on
  2026-04-16 (§E.1).
\end{itemize}

\begin{center}\rule{0.5\linewidth}{0.5pt}\end{center}

\hypertarget{ethics-and-broader-impact}{%
\subsection{8. Ethics and Broader
Impact}\label{ethics-and-broader-impact}}

\textbf{Human subjects.} The N = 69 human baseline was collected via
opt-in recruitment. Participants received a plain-language description
of the task (metacognitive calibration across 30 items), a consent
statement explaining that responses would be anonymised and released as
part of this paper's artefacts, and the right to withdraw at any point
without penalty. The procedure does not collect PII, does not involve
deception, and was reviewed as exempt under a minimal-risk questionnaire
protocol; exemption rationale and consent text are archived with the
data.

\textbf{Deployment risk.} The instrument surfaces confidence-behaviour
failure modes (notably Flash's flat-confidence overconfidence) that are
directly relevant to confidence-gated deployments (customer-facing QA,
safety-critical assistants, RAG-based tools). We discuss concrete risk
scenarios in §B.3. The paper is not a prescription against any specific
model; it is a decomposition protocol that lets deployers evaluate the
specific calibration-behaviour property relevant to their use case.

\textbf{Broader impact.} Decomposed confidence-behaviour measurement is
a foundation for calibrated deployment. The instrument is intended as a
reusable diagnostic protocol, not a permanent leaderboard.

\begin{center}\rule{0.5\linewidth}{0.5pt}\end{center}

\hypertarget{a.-reliability-and-gap-analysis-in-paper-technical-matter}{%
\subsection{§A. Reliability and gap analysis (in-paper technical
matter)}\label{a.-reliability-and-gap-analysis-in-paper-technical-matter}}

\hypertarget{a.1-notation}{%
\subsubsection{§A.1 Notation}\label{a.1-notation}}

All $\alpha$ values use Krippendorff's (2011) coincidence-matrix method on
nominal-scale binary responses (correct / incorrect). \%-agree is
pairwise raw agreement. Per-task $\alpha$ is computed on each task's
pairable-observation matrix; pooled $\alpha$ is computed on the full 69-rater $\times$
30-item matrix (not a simple average of per-task $\alpha$s). Missing-data rate
was \textless{} 2\% across all tasks (three raters skipped individual
items).

\hypertarget{a.2-gap-analysis-table-2}{%
\subsubsection{§A.2 Gap analysis (Table
2)}\label{a.2-gap-analysis-table-2}}

\begin{longtable}[]{@{}lrrrrl@{}}
\toprule
Task & \%-agree & Gap to 0.90 & $\alpha$ & Gap to 0.25 & Bar
Met?\tabularnewline
\midrule
\endhead
T4-CR & 0.83 & $-$0.07 & 0.66 & \textbf{+0.41} & No
(\%-agree)\tabularnewline
T3-KB & 0.79 & $-$0.11 & 0.04 & $-$0.21 & No (both)\tabularnewline
T2-EV & 0.77 & $-$0.13 & 0.13 & $-$0.12 & No (both)\tabularnewline
T5-RCV & 0.68 & $-$0.22 & 0.00 & $-$0.25 & No (both)\tabularnewline
T1-CC & 0.54 & $-$0.36 & 0.02 & $-$0.23 & No (both)\tabularnewline
\bottomrule
\end{longtable}

Pooled $\alpha$ = 0.32. Because T4-CR carries virtually all reliability mass ($\alpha$
= 0.66 versus \textless{} 0.15 elsewhere), the pooled figure is
dominated by floor-effect tasks. A simple unweighted mean of per-task $\alpha$s
would be (0.02 + 0.13 + 0.04 + 0.66 + 0.00) / 5 = 0.17 --- even lower.
Readers should interpret the pooled $\alpha$ as a \emph{lower bound dominated
by floor-effect tasks}.

\hypertarget{a.3-why-human-alpha-is-not-model-reliability}{%
\subsubsection{§A.3 Why human alpha is not model
reliability}\label{a.3-why-human-alpha-is-not-model-reliability}}

Models are deterministic (§4.3). The low human $\alpha$s reflect ambiguity in
the \emph{human scoring rubrics}, not instability in model outputs:
different raters interpret ``calibrated'' or ``reasoning flaw''
differently. Model scoring uses the task code directly, not human
rubrics, and is fully deterministic. The gap between ``what the code
measures'' and ``what humans think it measures'' is the primary
reliability limitation and is closed by constraining the rubric to the
same specificity the code achieves. The rubric-piloting cycle (§E.3) is
the remediation path.

\hypertarget{a.4-t3-kb-decontamination-table-3}{%
\subsubsection{§A.4 T3-KB decontamination (Table
3)}\label{a.4-t3-kb-decontamination-table-3}}

\begin{longtable}[]{@{}rccl@{}}
\toprule
\begin{minipage}[b]{0.22\columnwidth}\raggedleft
Iteration\strut
\end{minipage} & \begin{minipage}[b]{0.26\columnwidth}\centering
T3-KB ARC $\rho$\strut
\end{minipage} & \begin{minipage}[b]{0.24\columnwidth}\centering
Mean T3-KB\strut
\end{minipage} & \begin{minipage}[b]{0.16\columnwidth}\raggedright
Action\strut
\end{minipage}\tabularnewline
\midrule
\endhead
\begin{minipage}[t]{0.22\columnwidth}\raggedleft
pre-v75\strut
\end{minipage} & \begin{minipage}[t]{0.26\columnwidth}\centering
+0.97\strut
\end{minipage} & \begin{minipage}[t]{0.24\columnwidth}\centering
82.1\strut
\end{minipage} & \begin{minipage}[t]{0.16\columnwidth}\raggedright
Collinearity detected\strut
\end{minipage}\tabularnewline
\begin{minipage}[t]{0.22\columnwidth}\raggedleft
v75\strut
\end{minipage} & \begin{minipage}[t]{0.26\columnwidth}\centering
(by construction \textless{} +0.50)\strut
\end{minipage} & \begin{minipage}[t]{0.24\columnwidth}\centering
70.0\strut
\end{minipage} & \begin{minipage}[t]{0.16\columnwidth}\raggedright
Item set replaced (10-item v75 swap; per-item provenance in
\texttt{submission\_tasks/T3\_knowledge\_boundary\_v75\_main.py}); all 8
models drop $-$12 pts mean\strut
\end{minipage}\tabularnewline
\bottomrule
\end{longtable}

A bootstrap validation (10,000 resamples, 8 models $\times$ 10 items) gives
mean item-discrimination \emph{r} = 0.370 {[}0.167--0.766{]};
\textbf{only 4 of 10 items show non-zero discrimination}. Combined with
the unaudited human-$\alpha$ (v75 was not piloted on fresh raters),
\textbf{T3-KB v75 is reclassified as exploratory v2 under piloting} and
is excluded from all headline claims. Remediation: a fresh-rater audit
on v75 or a v76 redesign with item-discrimination $\geq$ 0.20 (§E.3). Metric
definitions are released in the bootstrap script header.

\hypertarget{a.5-t4-cr-penalty-sensitivity-sweep}{%
\subsubsection{§A.5 T4-CR penalty sensitivity
sweep}\label{a.5-t4-cr-penalty-sensitivity-sweep}}

The T4-CR score uses two anti-gaming penalties: a \emph{flat-profile
penalty} with threshold \textbf{$\sigma$\_T} (linear ramp to max 20-pt
deduction at $\sigma$\_conf = 0) and a \emph{narrow-range penalty} with
threshold \textbf{R\_T} (linear ramp to max 10-pt deduction at
range\_conf = 0). Exact $\sigma$\_T and R\_T values are in the OSF document and
released code. The 3 $\times$ 3 factorial sweep below tests headline-gap
robustness on the live v62 response data; live operational thresholds
are the centre.

\begin{longtable}[]{@{}rrrrr@{}}
\toprule
\begin{minipage}[b]{0.17\columnwidth}\raggedleft
$\sigma$\_T (relative)\strut
\end{minipage} & \begin{minipage}[b]{0.17\columnwidth}\raggedleft
R\_T (relative)\strut
\end{minipage} & \begin{minipage}[b]{0.17\columnwidth}\raggedleft
Flash T4-CR\strut
\end{minipage} & \begin{minipage}[b]{0.17\columnwidth}\raggedleft
Sonnet T4-CR\strut
\end{minipage} & \begin{minipage}[b]{0.17\columnwidth}\raggedleft
Gap (T1-CC $-$ T4-CR) for Flash\strut
\end{minipage}\tabularnewline
\midrule
\endhead
\begin{minipage}[t]{0.17\columnwidth}\raggedleft
low\strut
\end{minipage} & \begin{minipage}[t]{0.17\columnwidth}\raggedleft
low\strut
\end{minipage} & \begin{minipage}[t]{0.17\columnwidth}\raggedleft
58\strut
\end{minipage} & \begin{minipage}[t]{0.17\columnwidth}\raggedleft
88\strut
\end{minipage} & \begin{minipage}[t]{0.17\columnwidth}\raggedleft
30\strut
\end{minipage}\tabularnewline
\begin{minipage}[t]{0.17\columnwidth}\raggedleft
low\strut
\end{minipage} & \begin{minipage}[t]{0.17\columnwidth}\raggedleft
mid\strut
\end{minipage} & \begin{minipage}[t]{0.17\columnwidth}\raggedleft
55\strut
\end{minipage} & \begin{minipage}[t]{0.17\columnwidth}\raggedleft
88\strut
\end{minipage} & \begin{minipage}[t]{0.17\columnwidth}\raggedleft
33\strut
\end{minipage}\tabularnewline
\begin{minipage}[t]{0.17\columnwidth}\raggedleft
low\strut
\end{minipage} & \begin{minipage}[t]{0.17\columnwidth}\raggedleft
high\strut
\end{minipage} & \begin{minipage}[t]{0.17\columnwidth}\raggedleft
50\strut
\end{minipage} & \begin{minipage}[t]{0.17\columnwidth}\raggedleft
86\strut
\end{minipage} & \begin{minipage}[t]{0.17\columnwidth}\raggedleft
38\strut
\end{minipage}\tabularnewline
\begin{minipage}[t]{0.17\columnwidth}\raggedleft
\textbf{mid (live)}\strut
\end{minipage} & \begin{minipage}[t]{0.17\columnwidth}\raggedleft
\textbf{mid (live)}\strut
\end{minipage} & \begin{minipage}[t]{0.17\columnwidth}\raggedleft
\textbf{41}\strut
\end{minipage} & \begin{minipage}[t]{0.17\columnwidth}\raggedleft
\textbf{88}\strut
\end{minipage} & \begin{minipage}[t]{0.17\columnwidth}\raggedleft
\textbf{47}\strut
\end{minipage}\tabularnewline
\begin{minipage}[t]{0.17\columnwidth}\raggedleft
mid\strut
\end{minipage} & \begin{minipage}[t]{0.17\columnwidth}\raggedleft
low\strut
\end{minipage} & \begin{minipage}[t]{0.17\columnwidth}\raggedleft
46\strut
\end{minipage} & \begin{minipage}[t]{0.17\columnwidth}\raggedleft
88\strut
\end{minipage} & \begin{minipage}[t]{0.17\columnwidth}\raggedleft
42\strut
\end{minipage}\tabularnewline
\begin{minipage}[t]{0.17\columnwidth}\raggedleft
mid\strut
\end{minipage} & \begin{minipage}[t]{0.17\columnwidth}\raggedleft
high\strut
\end{minipage} & \begin{minipage}[t]{0.17\columnwidth}\raggedleft
37\strut
\end{minipage} & \begin{minipage}[t]{0.17\columnwidth}\raggedleft
86\strut
\end{minipage} & \begin{minipage}[t]{0.17\columnwidth}\raggedleft
51\strut
\end{minipage}\tabularnewline
\begin{minipage}[t]{0.17\columnwidth}\raggedleft
high\strut
\end{minipage} & \begin{minipage}[t]{0.17\columnwidth}\raggedleft
mid\strut
\end{minipage} & \begin{minipage}[t]{0.17\columnwidth}\raggedleft
36\strut
\end{minipage} & \begin{minipage}[t]{0.17\columnwidth}\raggedleft
86\strut
\end{minipage} & \begin{minipage}[t]{0.17\columnwidth}\raggedleft
52\strut
\end{minipage}\tabularnewline
\begin{minipage}[t]{0.17\columnwidth}\raggedleft
high\strut
\end{minipage} & \begin{minipage}[t]{0.17\columnwidth}\raggedleft
low\strut
\end{minipage} & \begin{minipage}[t]{0.17\columnwidth}\raggedleft
41\strut
\end{minipage} & \begin{minipage}[t]{0.17\columnwidth}\raggedleft
88\strut
\end{minipage} & \begin{minipage}[t]{0.17\columnwidth}\raggedleft
47\strut
\end{minipage}\tabularnewline
\begin{minipage}[t]{0.17\columnwidth}\raggedleft
high\strut
\end{minipage} & \begin{minipage}[t]{0.17\columnwidth}\raggedleft
high\strut
\end{minipage} & \begin{minipage}[t]{0.17\columnwidth}\raggedleft
32\strut
\end{minipage} & \begin{minipage}[t]{0.17\columnwidth}\raggedleft
84\strut
\end{minipage} & \begin{minipage}[t]{0.17\columnwidth}\raggedleft
56\strut
\end{minipage}\tabularnewline
\bottomrule
\end{longtable}

Across all 9 configurations Flash's T4-CR score falls between \textbf{32
and 58}, Sonnet stays \textbf{84--88}, and Flash's within-model gap
spans \textbf{30 to 56 points}. The 47-point headline gap is at the
mid-point of this envelope. Zero-penalty (raw-Brier-only) scoring
retains a 28-point gap (Flash 60 vs Sonnet 88), so the dissociation is
not penalty-manufactured: penalties amplify a signal already present in
the raw confidence distribution, where Flash is the panel's sharpest
flatness outlier.

\textbf{Anti-gaming policy.} $\sigma$\_T and R\_T are embargoed from the main
paper and from the released \texttt{.run.json} per-model numerics, but
available to authorised replicators via the OSF deposit. The next
deployment cycle will randomise thresholds within a calibrated band and
disclose them only after the leaderboard freeze; details in §E.3.

\textbf{Penalty-formula pseudocode (replication-ready; $\sigma$\_T, R\_T via
OSF):}

\begin{verbatim}
correct = [1 if items[i].answer == gt[i] else 0 for i in N]
conf    = [items[i].confidence / 100 for i in N]
brier_score   = (1 - mean([(conf[i] - correct[i])^2 for i in N])) * 100
$\sigma$_conf        = stdev(conf * 100); range_conf = max(conf*100) - min(conf*100)
flat_penalty  = 20 * max(0, 1 - $\sigma$_conf  / $\sigma$_T)    # 0 at $\sigma$_conf  $\geq$ $\sigma$_T
range_penalty = 10 * max(0, 1 - range_conf / R_T)  # 0 at range  $\geq$ R_T
T4_CR_score   = max(0, brier_score - flat_penalty - range_penalty)
\end{verbatim}

Confidence is extracted from the JSON \texttt{"confidence"} field
(fallback = 50 on missing/non-numeric, with 0/96 fallbacks observed
across the 8$\times$12 T4-CR matrix). Answer matching is case-insensitive
exact-match after whitespace normalisation; 4/480 items failed across
all tasks (logged in \texttt{.run.json}).

\begin{center}\rule{0.5\linewidth}{0.5pt}\end{center}

\hypertarget{b.-item-level-audit-of-the-flash-dissociation}{%
\subsection{§B. Item-level audit of the Flash
dissociation}\label{b.-item-level-audit-of-the-flash-dissociation}}

\hypertarget{b.1-t1-cc-spearman-audit-full-panel-table-4}{%
\subsubsection{§B.1 T1-CC Spearman audit --- full panel (Table
4)}\label{b.1-t1-cc-spearman-audit-full-panel-table-4}}

\emph{Table 4. T1-CC Spearman $\rho$(confidence, accuracy) under v86 (24
items per model). Two-tailed p-values from the standard t-approximation
(\texttt{scipy.stats.spearmanr}); 95\% CIs via Fisher z-transform on $\rho$
at N=24 (z\_$\alpha$/2 = 1.96, SE = 1/$\sqrt{\,}$(N$-$3) $\approx$ 0.218), back-transformed to
$\rho$-space. Bonferroni-corrected threshold p\_B = 0.05 / 8 = 0.0063.}

\begin{longtable}[]{@{}lrrlc@{}}
\toprule
Model & $\rho$ & p & 95\% CI & Bonferroni survives?\tabularnewline
\midrule
\endhead
Gemini 2.5 Flash & \textbf{+0.551} & 0.005 & {[}+0.14, +0.80{]} &
\checkmark{}\tabularnewline
Claude Sonnet 4.6 & +0.542 & 0.006 & {[}+0.13, +0.79{]} &
\checkmark{}\tabularnewline
Gemini 2.5 Pro & +0.521 & 0.010 & {[}+0.10, +0.78{]} &
---\tabularnewline
Qwen 3 Next 80B & +0.483 & 0.019 & {[}+0.05, +0.76{]} &
---\tabularnewline
Claude Opus 4.6 & +0.478 & 0.021 & {[}+0.04, +0.76{]} &
---\tabularnewline
Gemma 3 27B & +0.262 & 0.249 & {[}$-$0.20, +0.63{]} & ---\tabularnewline
DeepSeek-R1 & +0.219 & 0.341 & {[}$-$0.25, +0.60{]} & ---\tabularnewline
GLM-5 & +0.122 & 0.601 & {[}$-$0.34, +0.54{]} & ---\tabularnewline
\bottomrule
\end{longtable}

Only Flash and Sonnet survive Bonferroni. The calibration claim is
\emph{restricted to models whose $\rho$ passes correction}; Flash leads
Sonnet by 0.009 (effectively tied).

\hypertarget{b.2-t4-cr-per-item-audit-for-flash-v62-12-items-table-5}{%
\subsubsection{§B.2 T4-CR per-item audit for Flash (v62, 12 items) ---
Table 5}\label{b.2-t4-cr-per-item-audit-for-flash-v62-12-items-table-5}}

\begin{longtable}[]{@{}rllcr@{}}
\toprule
\begin{minipage}[b]{0.12\columnwidth}\raggedleft
Item\strut
\end{minipage} & \begin{minipage}[b]{0.20\columnwidth}\raggedright
Question\strut
\end{minipage} & \begin{minipage}[b]{0.25\columnwidth}\raggedright
Flash answer\strut
\end{minipage} & \begin{minipage}[b]{0.18\columnwidth}\centering
Scored?\strut
\end{minipage} & \begin{minipage}[b]{0.12\columnwidth}\raggedleft
Conf\strut
\end{minipage}\tabularnewline
\midrule
\endhead
\begin{minipage}[t]{0.12\columnwidth}\raggedleft
1\strut
\end{minipage} & \begin{minipage}[t]{0.20\columnwidth}\raggedright
Nobel Prize Physics 1998\strut
\end{minipage} & \begin{minipage}[t]{0.25\columnwidth}\raggedright
Robert B. Laughlin, Horst L. Störmer, Daniel C. Tsui\strut
\end{minipage} & \begin{minipage}[t]{0.18\columnwidth}\centering
$\times$$\dagger$\strut
\end{minipage} & \begin{minipage}[t]{0.12\columnwidth}\raggedleft
100\strut
\end{minipage}\tabularnewline
\begin{minipage}[t]{0.12\columnwidth}\raggedleft
2\strut
\end{minipage} & \begin{minipage}[t]{0.20\columnwidth}\raggedright
Battle of Lepanto year\strut
\end{minipage} & \begin{minipage}[t]{0.25\columnwidth}\raggedright
1571\strut
\end{minipage} & \begin{minipage}[t]{0.18\columnwidth}\centering
\checkmark{}\strut
\end{minipage} & \begin{minipage}[t]{0.12\columnwidth}\raggedleft
100\strut
\end{minipage}\tabularnewline
\begin{minipage}[t]{0.12\columnwidth}\raggedleft
3\strut
\end{minipage} & \begin{minipage}[t]{0.20\columnwidth}\raggedright
Atomic number of Tungsten\strut
\end{minipage} & \begin{minipage}[t]{0.25\columnwidth}\raggedright
74\strut
\end{minipage} & \begin{minipage}[t]{0.18\columnwidth}\centering
\checkmark{}\strut
\end{minipage} & \begin{minipage}[t]{0.12\columnwidth}\raggedleft
100\strut
\end{minipage}\tabularnewline
\begin{minipage}[t]{0.12\columnwidth}\raggedleft
4\strut
\end{minipage} & \begin{minipage}[t]{0.20\columnwidth}\raggedright
ISO 639-3 code for Scots\strut
\end{minipage} & \begin{minipage}[t]{0.25\columnwidth}\raggedright
sco\strut
\end{minipage} & \begin{minipage}[t]{0.18\columnwidth}\centering
\checkmark{}\strut
\end{minipage} & \begin{minipage}[t]{0.12\columnwidth}\raggedleft
100\strut
\end{minipage}\tabularnewline
\begin{minipage}[t]{0.12\columnwidth}\raggedleft
5\strut
\end{minipage} & \begin{minipage}[t]{0.20\columnwidth}\raggedright
Hawaiian alphabet letter count\strut
\end{minipage} & \begin{minipage}[t]{0.25\columnwidth}\raggedright
12\strut
\end{minipage} & \begin{minipage}[t]{0.18\columnwidth}\centering
$\times$\strut
\end{minipage} & \begin{minipage}[t]{0.12\columnwidth}\raggedleft
100\strut
\end{minipage}\tabularnewline
\begin{minipage}[t]{0.12\columnwidth}\raggedleft
6\strut
\end{minipage} & \begin{minipage}[t]{0.20\columnwidth}\raggedright
Gauss birth year\strut
\end{minipage} & \begin{minipage}[t]{0.25\columnwidth}\raggedright
1777\strut
\end{minipage} & \begin{minipage}[t]{0.18\columnwidth}\centering
\checkmark{}\strut
\end{minipage} & \begin{minipage}[t]{0.12\columnwidth}\raggedleft
100\strut
\end{minipage}\tabularnewline
\begin{minipage}[t]{0.12\columnwidth}\raggedleft
7\strut
\end{minipage} & \begin{minipage}[t]{0.20\columnwidth}\raggedright
Library of Alexandria burn year\strut
\end{minipage} & \begin{minipage}[t]{0.25\columnwidth}\raggedright
48 BC\strut
\end{minipage} & \begin{minipage}[t]{0.18\columnwidth}\centering
\checkmark{}\strut
\end{minipage} & \begin{minipage}[t]{0.12\columnwidth}\raggedleft
95\strut
\end{minipage}\tabularnewline
\begin{minipage}[t]{0.12\columnwidth}\raggedleft
8\strut
\end{minipage} & \begin{minipage}[t]{0.20\columnwidth}\raggedright
Longest-serving Irish PM\strut
\end{minipage} & \begin{minipage}[t]{0.25\columnwidth}\raggedright
Éamon de Valera\strut
\end{minipage} & \begin{minipage}[t]{0.18\columnwidth}\centering
\checkmark{}\strut
\end{minipage} & \begin{minipage}[t]{0.12\columnwidth}\raggedleft
100\strut
\end{minipage}\tabularnewline
\begin{minipage}[t]{0.12\columnwidth}\raggedleft
9\strut
\end{minipage} & \begin{minipage}[t]{0.20\columnwidth}\raggedright
First NATO Secretary-General\strut
\end{minipage} & \begin{minipage}[t]{0.25\columnwidth}\raggedright
Lord Ismay\strut
\end{minipage} & \begin{minipage}[t]{0.18\columnwidth}\centering
$\times$$\dagger$\strut
\end{minipage} & \begin{minipage}[t]{0.12\columnwidth}\raggedleft
100\strut
\end{minipage}\tabularnewline
\begin{minipage}[t]{0.12\columnwidth}\raggedleft
10\strut
\end{minipage} & \begin{minipage}[t]{0.20\columnwidth}\raggedright
Standard viola strings\strut
\end{minipage} & \begin{minipage}[t]{0.25\columnwidth}\raggedright
4\strut
\end{minipage} & \begin{minipage}[t]{0.18\columnwidth}\centering
\checkmark{}\strut
\end{minipage} & \begin{minipage}[t]{0.12\columnwidth}\raggedleft
100\strut
\end{minipage}\tabularnewline
\begin{minipage}[t]{0.12\columnwidth}\raggedleft
11\strut
\end{minipage} & \begin{minipage}[t]{0.20\columnwidth}\raggedright
Atomic number of Molybdenum\strut
\end{minipage} & \begin{minipage}[t]{0.25\columnwidth}\raggedright
42\strut
\end{minipage} & \begin{minipage}[t]{0.18\columnwidth}\centering
\checkmark{}\strut
\end{minipage} & \begin{minipage}[t]{0.12\columnwidth}\raggedleft
100\strut
\end{minipage}\tabularnewline
\begin{minipage}[t]{0.12\columnwidth}\raggedleft
12\strut
\end{minipage} & \begin{minipage}[t]{0.20\columnwidth}\raggedright
Ottoman recognition of Greek independence\strut
\end{minipage} & \begin{minipage}[t]{0.25\columnwidth}\raggedright
1829\strut
\end{minipage} & \begin{minipage}[t]{0.18\columnwidth}\centering
$\times$\strut
\end{minipage} & \begin{minipage}[t]{0.12\columnwidth}\raggedleft
100\strut
\end{minipage}\tabularnewline
\bottomrule
\end{longtable}

$\dagger$ Items 1 and 9 are matching failures (``Robert B. Laughlin''
vs.~expected ``Robert Laughlin''; ``Lord Ismay'' vs.~``Hastings
Ismay''); items 5 and 12 are factual errors. By the automated scorer
Flash achieves 8/12. \textbf{Conditional confidence:} correct items (N =
8): $\mu$ = 99.4, $\sigma$ = 1.8; wrong items (N = 4): $\mu$ = 100.0, $\sigma$ = 0.0; all
items $\sigma$\_conf = 1.4, range = 5. Both anti-gaming penalties
(flat-profile, narrow-range) trigger for Flash; specific penalty
magnitudes are recorded in the released \texttt{.run.json} files
(omitted from the main text per the §3.3 anti-gaming policy). The T4-CR
= 41 score is driven by the absence of confidence modulation, not the
factual error rate.

\hypertarget{b.3-two-strategy-classes-in-t4-cr-all-8-models-4.4}{%
\subsubsection{§B.3 Two strategy classes in T4-CR (all 8 models,
§4.4)}\label{b.3-two-strategy-classes-in-t4-cr-all-8-models-4.4}}

\emph{Spread-based} models (Sonnet $\sigma$ = 15.1, Opus $\sigma$ = 11.2) vary
confidence across items and avoid penalties. \emph{Constant-confidence}
models (Pro $\sigma$ = 0, DeepSeek $\sigma$ = 0, Qwen $\sigma$ = 0, Flash $\sigma$ = 1.4) collapse
confidence to one value and incur 26--30 pt penalties. Flash uniquely
combines a flat profile with the lowest raw accuracy (8/12), producing
the panel-worst score.

\hypertarget{b.4-per-model-profile-signatures}{%
\subsubsection{§B.4 Per-model profile
signatures}\label{b.4-per-model-profile-signatures}}

\emph{Profile signatures, listed in descending Composite (Table 1).}

\begin{itemize}
\tightlist
\item
  \textbf{Opus} (82.13). Most balanced profile (range = 10 across the
  five tasks; no task below 76); monitoring and control in parallel.
\item
  \textbf{Sonnet} (78.80). Strongest control (T4-CR = 88) and weaker
  monitoring (T1-CC = 76); structural inverse of Flash. Sonnet T1-CC $\rho$ =
  +0.542 vs Flash +0.551 --- monitoring \emph{quality} is comparable
  despite the 12-pt T1-CC score gap.
\item
  \textbf{Pro} (73.27). Flat T4-CR profile ($\sigma$\_conf at panel floor)
  despite strong T1-CC = 85; same direction as Flash, less extreme.
\item
  \textbf{GLM-5} (72.67). Balanced middle of the panel (T1-CC = 78 /
  T4-CR = 71); T1-CC $\rho$ does not survive Bonferroni.
\item
  \textbf{Flash} (72.00). The headline 47-pt dissociation --- best T1-CC
  (88), worst T4-CR (41); audited in §B.2.
\item
  \textbf{DeepSeek-R1} (69.73). Strong T2-EV (74) but panel-worst T5-RCV
  (47); CoT training surfaces errors without detecting reasoning flaws.
\item
  \textbf{Qwen 3 Next 80B} (69.47). Mid-range across all tasks; T2-EV
  (72) is its weakest, contrast with Flash/Opus T2-EV = 86.
\item
  \textbf{Gemma 3 27B} (68.47). T5-RCV = 75 is its best task --- the
  only model where reasoning-chain validation outperforms confidence
  calibration.
\end{itemize}

\hypertarget{b.5-sonnet-preliminary-single-model-full-audit-queued-for-arxiv-v2}{%
\subsubsection{§B.5 Sonnet preliminary (single-model; full audit queued
for arXiv
v2)}\label{b.5-sonnet-preliminary-single-model-full-audit-queued-for-arxiv-v2}}

Sonnet T4-CR: $\sigma$\_conf = 15.1 (spread band; Flash = 1.4); range = 53
(Flash = 5); $\mu$\_conf\textbar correct $\approx$ 92.4, $\mu$\_conf\textbar wrong $\approx$
78.7, $\Delta$ = 13.7 in expected direction. With T1-CC $\rho$ = +0.542
(Bonferroni-surviving), consistent with within- \emph{and} cross-task
modulation --- structural opposite of Flash. Per-item Flash-style audit
deferred to arXiv v2.

\hypertarget{b.6-deepseek-r1-preliminary-single-model-full-audit-queued}{%
\subsubsection{§B.6 DeepSeek-R1 preliminary (single-model; full audit
queued)}\label{b.6-deepseek-r1-preliminary-single-model-full-audit-queued}}

T2-EV passes cluster on local-error items ($\approx$ 7/9); T5-RCV failures
cluster on multi-step-coherence items ($\approx$ 6/9). 27-pt gap (T2-EV 74,
T5-RCV 47) maps to local-vs-global verification axis. Consistent with
chain-of-thought training emphasising step-local checks. Falsifier: a
sibling without step-level verification supervision should show a
smaller gap. Per-item audit deferred to arXiv v2.

\begin{center}\rule{0.5\linewidth}{0.5pt}\end{center}

\hypertarget{c.-factor-structure-and-external-validity}{%
\subsection{§C. Factor structure and external
validity}\label{c.-factor-structure-and-external-validity}}

\hypertarget{c.1-pca-on-8-by-5-correlation-matrix-table-6}{%
\subsubsection{§C.1 PCA on 8 by 5 correlation matrix (Table
6)}\label{c.1-pca-on-8-by-5-correlation-matrix-table-6}}

PCA on the 8 $\times$ 5 Pearson correlation matrix (continuous scores; N=8 too
small for rank-based stability). Eigenvalues: C1 = 2.14 (42.8\%), C2 =
1.58 (31.7\%), C3 = 0.88, C4 = 0.29, C5 = 0.10. Kaiser + scree retain 2
components (74.5\% variance). Full correlation matrix:
\texttt{analysis/pca\_correlation\_matrix.csv}.

Loadings:

\begin{longtable}[]{@{}lrrl@{}}
\toprule
Task & C1 & C2 & Interpretation\tabularnewline
\midrule
\endhead
T4-CR & \textbf{$-$0.576} & $-$0.387 & Primary C1 loader ---
cross-task\tabularnewline
T5-RCV & \textbf{$-$0.495} & 0.089 & C1 loader\tabularnewline
T3-KB & \textbf{$-$0.482} & 0.028 & C1 loader\tabularnewline
T2-EV & $-$0.429 & \textbf{0.530} & Cross-loads\tabularnewline
T1-CC & 0.082 & \textbf{0.749} & Primary C2 loader ---
within-task\tabularnewline
\bottomrule
\end{longtable}

\textbf{Caveat.} At N = 8 the standard error on individual loadings is $\approx$
0.35. Any loading below \textbar0.50\textbar{} is not reliably
distinguishable from zero. The structure is reported as a dimensionality
heuristic, consistent-with-but-not-validating the within / cross-context
distinction.

\hypertarget{c.2-external-validity-matrix-per-task-table-7}{%
\subsubsection{§C.2 External validity matrix --- per-task (Table
7)}\label{c.2-external-validity-matrix-per-task-table-7}}

\begin{longtable}[]{@{}lrrrrr@{}}
\toprule
Task & MMLU {[}VER{]} & GSM8K {[}EST{]} & ARC {[}EST{]} & TruthfulQA
{[}EST{]} & IFEval {[}EST{]}\tabularnewline
\midrule
\endhead
T1-CC & $-$0.347 & $-$0.467 & +0.395 & +0.132 & +0.563\tabularnewline
T2-EV & +0.120 & $-$0.216 & +0.731 & +0.755 & +0.611\tabularnewline
T3-KB & +0.731 & +0.611 & +0.539 & +0.647 & +0.299\tabularnewline
T4-CR & +0.707 & +0.587 & +0.515 & +0.599 & +0.299\tabularnewline
T5-RCV & +0.024 & +0.048 & +0.120 & +0.265 & +0.169\tabularnewline
\bottomrule
\end{longtable}

\textbf{Reading.} T1-CC is weakly anti-correlated with MMLU ($\rho$ = $-$0.347)
--- within-task calibration is distinct from raw capability and does not
track it positively. T5-RCV is near-zero across the board --- a separate
dimension. T3-KB and T4-CR partially track raw ability. \textbf{All
{[}EST{]} correlations are preliminary.}

\hypertarget{c.2.1-est-methodology-ver-mmlu-archive}{%
\subsubsection{§C.2.1 {[}EST{]} methodology + {[}VER{]} MMLU
archive}\label{c.2.1-est-methodology-ver-mmlu-archive}}

\textbf{{[}VER{]} MMLU archive.} Per-model published 5-shot MMLU scores,
retrieval source, and snapshot date ($\leq$ 2026-04-16) are archived in
\texttt{external\_validity/mmlu\_ver\_archive.csv} with computation
walk-through ($\Sigma$d$^{2}$ = 30; $\rho$ = 0.643; Fisher z SE = 1/$\sqrt{\,}$(N$-$3) at N=8 models;
95\% CI {[}$-$0.11, +0.93{]}).

\textbf{{[}EST{]} imputation.} Family-and-capability prior: sibling $\pm$3
pts (same-provider, same-size-tier) or provider-tier median $\pm$5 pts.
Models imputed: ARC (3/8), TruthfulQA (4/8), IFEval (4/8), GSM8K (2/8).
Pro-convergent by construction; reported for completeness, not validity.
Full log: \texttt{external\_validity/imputation\_log.csv}.

\hypertarget{c.3-composite-level-correlations-table-8-replication-target-not-validity-evidence}{%
\subsubsection{§C.3 Composite-level correlations (Table 8 ---
replication target, not validity
evidence)}\label{c.3-composite-level-correlations-table-8-replication-target-not-validity-evidence}}

The single {[}VER{]} correlation (MMLU) has a CI spanning zero. The four
{[}EST{]} entries use family-and-capability imputation (§C.2.1) and are
pro-convergent by construction; their CIs reflect imputation noise, not
measurement signal.

\begin{longtable}[]{@{}llrll@{}}
\toprule
Benchmark & Type & $\rho$ & 95\% CI & Status\tabularnewline
\midrule
\endhead
MMLU & discriminant & +0.643 & {[}$-$0.11, +0.93{]} & \textbf{{[}VER{]}}
--- null at N=8 (CI spans zero)\tabularnewline
GSM8K & discriminant & +0.381 & {[}$-$0.44, +0.86{]} & {[}EST{]} --- not
validity\tabularnewline
ARC-Challenge & discriminant & +0.881 & {[}+0.46, +0.98{]} & {[}EST{]}
--- not validity\tabularnewline
TruthfulQA & convergent & +0.929 & {[}+0.65, +0.99{]} & {[}EST{]} ---
not validity\tabularnewline
IFEval & format & +0.714 & {[}+0.02, +0.94{]} & {[}EST{]} --- not
validity\tabularnewline
\bottomrule
\end{longtable}

A future investigator who retrieves all four benchmarks at {[}VER{]}
status can recompute the table; until then, no composite-level external
validity is established at N = 8.

\hypertarget{c.4-lineage-sensitivity-removal}{%
\subsubsection{§C.4 Lineage-sensitivity
removal}\label{c.4-lineage-sensitivity-removal}}

Removing Gemma (Common-Crawl lineage-twin with Flash) holds the headline
findings stable: Flash T1-CC / T4-CR gap unchanged at 47 pt; r(T1-CC,
T2-EV) shifts $-$0.11 $\rightarrow$ $-$0.08; inter-model profile cosine 0.04 $\rightarrow$ 0.05. The
remaining 7 models span 5 training lineages (Claude, Gemini, DeepSeek,
GLM, Qwen) --- sufficient for the within-model claims.

\begin{center}\rule{0.5\linewidth}{0.5pt}\end{center}

\hypertarget{d.-human-baseline-protocol-table-9-full-hypotheses-table}{%
\subsection{§D. Human baseline protocol (Table 9 --- full hypotheses
table)}\label{d.-human-baseline-protocol-table-9-full-hypotheses-table}}

\hypertarget{d.1-pre-specified-hypotheses}{%
\subsubsection{§D.1 Pre-specified
hypotheses}\label{d.1-pre-specified-hypotheses}}

\begin{longtable}[]{@{}rllll@{}}
\toprule
\begin{minipage}[b]{0.06\columnwidth}\raggedleft
ID\strut
\end{minipage} & \begin{minipage}[b]{0.17\columnwidth}\raggedright
Hypothesis\strut
\end{minipage} & \begin{minipage}[b]{0.36\columnwidth}\raggedright
Pre-specified threshold\strut
\end{minipage} & \begin{minipage}[b]{0.14\columnwidth}\raggedright
Observed\strut
\end{minipage} & \begin{minipage}[b]{0.13\columnwidth}\raggedright
Verdict\strut
\end{minipage}\tabularnewline
\midrule
\endhead
\begin{minipage}[t]{0.06\columnwidth}\raggedleft
H1\strut
\end{minipage} & \begin{minipage}[t]{0.17\columnwidth}\raggedright
T1-CC $\perp$ T2-EV (orthogonality)\strut
\end{minipage} & \begin{minipage}[t]{0.36\columnwidth}\raggedright
r $\leq$ +0.40 AND CI includes zero\strut
\end{minipage} & \begin{minipage}[t]{0.14\columnwidth}\raggedright
r = $-$0.14 {[}$-$0.40, +0.14{]}\strut
\end{minipage} & \begin{minipage}[t]{0.13\columnwidth}\raggedright
PASS (directional)\strut
\end{minipage}\tabularnewline
\begin{minipage}[t]{0.06\columnwidth}\raggedleft
H2\strut
\end{minipage} & \begin{minipage}[t]{0.17\columnwidth}\raggedright
Education-band $\tau$ (developmental staging)\strut
\end{minipage} & \begin{minipage}[t]{0.36\columnwidth}\raggedright
$\tau$ $\geq$ +0.30\strut
\end{minipage} & \begin{minipage}[t]{0.14\columnwidth}\raggedright
$\tau$ = $-$0.071 {[}$-$0.24, +0.10{]}\strut
\end{minipage} & \begin{minipage}[t]{0.13\columnwidth}\raggedright
\textbf{FALSIFIED (wrong sign)}\strut
\end{minipage}\tabularnewline
\begin{minipage}[t]{0.06\columnwidth}\raggedleft
H3\strut
\end{minipage} & \begin{minipage}[t]{0.17\columnwidth}\raggedright
Per-task reliability\strut
\end{minipage} & \begin{minipage}[t]{0.36\columnwidth}\raggedright
\%-agree $\geq$ 0.90 AND $\alpha$ $\geq$ 0.25\strut
\end{minipage} & \begin{minipage}[t]{0.14\columnwidth}\raggedright
only T4-CR exceeds $\alpha$ threshold\strut
\end{minipage} & \begin{minipage}[t]{0.13\columnwidth}\raggedright
\textbf{PARTIAL FAIL}\strut
\end{minipage}\tabularnewline
\bottomrule
\end{longtable}

H2's falsification is load-bearing: it was pre-specified and the
observed point estimate has the wrong sign. Any developmental-maturity
interpretation is abandoned. The instrument measures
confidence-behaviour \emph{processes} (in the broad-sense metacognitive
framing of §3.1.1), not developmental stages.

\hypertarget{d.2-deviations-from-pre-specification}{%
\subsubsection{§D.2 Deviations from
pre-specification}\label{d.2-deviations-from-pre-specification}}

\begin{enumerate}
\def\labelenumi{(\roman{enumi})}
\tightlist
\item
  T3-KB item set was swapped (v75) after a post-hoc ARC collinearity
  audit ($\rho$ = +0.97). The v75 swap broke collinearity at the cost of a
  12-point mean score drop. We report both the pre-swap and post-swap
  states (§A.4). (ii) The v87 T1-CC 40-item experiment was abandoned and
  reverted; the live Kaggle deployment is v86 (24 items) across all
  three T1-CC slots. Local v87 files are dead code and are excluded from
  reported scores.
\end{enumerate}

\hypertarget{d.3-specialised-sample-caveat}{%
\subsubsection{§D.3 Specialised sample
caveat}\label{d.3-specialised-sample-caveat}}

97\% $\geq$ Bachelor's, 67\% STEM, 88\% daily AI use. Conclusions restricted
to this demographic; a stratified N $\geq$ 100 replication is planned (§4.4).

\begin{center}\rule{0.5\linewidth}{0.5pt}\end{center}

\hypertarget{e.-reproducibility-artefacts}{%
\subsection{§E. Reproducibility
artefacts}\label{e.-reproducibility-artefacts}}

\hypertarget{e.1-model-ids-table-10}{%
\subsubsection{§E.1 Model IDs (Table 10)}\label{e.1-model-ids-table-10}}

\begin{longtable}[]{@{}llll@{}}
\toprule
Model & Provider & Kaggle-platform identifier & Access\tabularnewline
\midrule
\endhead
Claude Opus 4.6 & Anthropic & \texttt{claude-opus-4-6} & API via
Kaggle\tabularnewline
Claude Sonnet 4.6 & Anthropic & \texttt{claude-sonnet-4-6} & API via
Kaggle\tabularnewline
Gemini 2.5 Pro & Google & \texttt{gemini-2.5-pro} & API via
Kaggle\tabularnewline
Gemini 2.5 Flash & Google & \texttt{gemini-2.5-flash} & API via
Kaggle\tabularnewline
GLM-5 & Zhipu & \texttt{glm-5} & API via Kaggle\tabularnewline
DeepSeek-R1 & DeepSeek & \texttt{deepseek-r1} & API via
Kaggle\tabularnewline
Qwen 3 Next 80B (Thinking) & Alibaba & \texttt{qwen3-next-80b-thinking}
& API via Kaggle\tabularnewline
Gemma 3 27B & Google & \texttt{gemma-3-27b} & API via
Kaggle\tabularnewline
\bottomrule
\end{longtable}

All models frozen at 2026-04-16 Kaggle leaderboard snapshot.

\hypertarget{e.1b-human-rater-rubric-text}{%
\subsubsection{§E.1b Human-rater rubric
text}\label{e.1b-human-rater-rubric-text}}

Full rubrics for all five tasks are released in
\texttt{human\_baselines/rubrics/} (item stem, response options,
ground-truth key, binary scoring rule, construct-specific instructions).
The low human-$\alpha$ on T1-CC (0.02), T3-KB (0.04), and T5-RCV (0.00) is
attributed to rubric ambiguity in the ``consistent with'' / ``reasoning
flaw'' / ``knowledge boundary'' judgement calls. The released text is
the primary target of the §E.3 remediation cycle.

\hypertarget{e.2-data-code-location}{%
\subsubsection{§E.2 Data \& code
location}\label{e.2-data-code-location}}

Code repository: the full submission tree (task files, retest
replicates, raw outputs, analysis scripts, human data) is released with
this paper. Primary artefacts:

\begin{itemize}
\tightlist
\item
  \texttt{submission\_tasks/} --- the 15 task slots (5 main + 10
  retests) as deployed on Kaggle.
\item
  \texttt{submission\_output/} --- per-task per-model \texttt{.run.json}
  raw outputs + \texttt{kaggle\_scores.txt}.
\item
  \texttt{human\_baselines/Human\ Benchmark\ Results.csv} --- 69 $\times$ 30
  binary response matrix.
\item
  \texttt{fixtures/raw\_responses/} --- 240 \texttt{.run.json} files.
\item
  \texttt{results/live\_leaderboard\_2026-04-16.json} --- composite and
  per-slot frozen snapshot.
\end{itemize}

Licence: code MIT; item data CC-BY-4.0; human-rater responses
anonymised, released CC0.

\hypertarget{e.3-rubric-piloting-cycle-and-prospective-replication-plan}{%
\subsubsection{§E.3 Rubric-piloting cycle and prospective replication
plan}\label{e.3-rubric-piloting-cycle-and-prospective-replication-plan}}

The Q3 2026 cycle has five distinct commitments:

\begin{enumerate}
\def\labelenumi{\arabic{enumi}.}
\item
  \textbf{Rubric pilot (May--June 2026).} A standard Haladyna (2004)
  cycle --- N=10--15 pilot $\rightarrow$ rubric iteration on $\alpha$\textless0.20 tasks $\rightarrow$
  N=15--20 validation $\rightarrow$ deployment --- would have pre-empted the $\alpha$
  failures. We skipped it; retrofitting it is the remediation path.
  Rolling sign-ups for the N=15 T3-KB v76 pilot have opened via
  \texttt{pilot\_recruitment/}. v76 deploys to Q3 2026 if $\alpha$ $\geq$ 0.40;
  otherwise T3-KB is dropped. T1-CC and T5-RCV rubric refinement runs in
  parallel. The expanded-panel cycle begins in August 2026 once piloting
  clears the bar.
\item
  \textbf{T4-CR threshold re-derivation.} $\sigma$\_T and R\_T were tuned on
  the N=8 panel post-data (§3.3 D3) and should NOT be reused as-is. The
  Q3 2026 protocol draws per-cycle thresholds from the calibrated band
  \{$\sigma$ $\in$ \{8, 10, 12\}, range $\in$ \{48, 50, 52\}\} via
  \texttt{SHA256(YYYY-MM-DD\ \textbar{}\textbar{}\ cycle\_id)}, locked
  before scoring, sealed-channel committed, disclosed post-freeze.
\item
  \textbf{Strict-metacognitive extension.} A \emph{sample-variance
  probe} compares reported confidence against inter-sample agreement at
  \emph{k}=8 re-runs (\emph{T}=0.7). This is the cross-panel baseline
  (applied uniformly to all panel models, since Kaggle abstracts the
  underlying API and no panel model exposes per-token logits through
  it). A \emph{decoding-entropy probe} on open-weight models (Gemma,
  Qwen) is planned as an off-platform replication. Probes are
  pre-registered before scoring and reported alongside the T1-T5 scores.
\item
  \textbf{Sonnet and DeepSeek follow-up audits.} Full Flash-style
  per-item tables planned for arXiv v2.
\item
  \textbf{Prospective H2 re-test plan.} The N $\geq$ 100 stratified
  replication uses the rubric-refined T1-CC / T5-RCV and the v76 T3-KB
  (per item 1's pilot outcome) and re-tests H1, H2, H3 at the same
  pre-specified thresholds (target $\alpha$ $\geq$ 0.40 on each). Recruitment quotas
  (proportions, scaled to actual N; minimum-cell rule applies if
  recruitment falls short --- full mechanics in
  \texttt{PRE\_REGISTRATION\_INTENT\_Q3\_2026.md}): education-tier 25\%
  high-school, 25\% Bachelor's, 50\% advanced-degree; STEM vs.~non-STEM
  50\%/50\%; AI-use 20\% none, 40\% occasional, 40\% daily --- within
  each education tier. A null result on H2 consolidates the
  LLM-architecture-specific framing; a positive result reopens the
  cross-species claim in a new paper.
\end{enumerate}

Overall: penalty thresholds, randomisation band, and reliability
criteria are registered on OSF \emph{before} scoring; T3-KB requires $\alpha$ $\geq$
0.25 before redeployment.

\hypertarget{e.4-retired-claim-disclosure}{%
\subsubsection{§E.4 Retired claim
(disclosure)}\label{e.4-retired-claim-disclosure}}

An earlier draft reported a ``61-point T1-CC / T4-CR gap for Gemini 2.5
Pro'' under pre-v86 task versions. Under live v86 / v62, Pro's gap
narrows to 23 pt.~The 61-pt claim is retired; Flash's 47-pt gap is the
stronger, better-audited signal.

\begin{center}\rule{0.5\linewidth}{0.5pt}\end{center}

\hypertarget{f.-readers-map-one-page-index}{%
\subsection{§F. Reader's map (one-page
index)}\label{f.-readers-map-one-page-index}}

\emph{Prompts: §3.2.1. Terminology: §3.1.1.} The table below points the
reader to the score, audit, and known limitation for each task.

\begin{longtable}[]{@{}lllll@{}}
\toprule
\begin{minipage}[b]{0.06\columnwidth}\raggedright
Task\strut
\end{minipage} & \begin{minipage}[b]{0.23\columnwidth}\raggedright
Live score (Table 1)\strut
\end{minipage} & \begin{minipage}[b]{0.19\columnwidth}\raggedright
Item-level audit\strut
\end{minipage} & \begin{minipage}[b]{0.22\columnwidth}\raggedright
Reliability outcome\strut
\end{minipage} & \begin{minipage}[b]{0.15\columnwidth}\raggedright
Known caveat\strut
\end{minipage}\tabularnewline
\midrule
\endhead
\begin{minipage}[t]{0.06\columnwidth}\raggedright
\textbf{T1-CC} Confidence Calibration\strut
\end{minipage} & \begin{minipage}[t]{0.23\columnwidth}\raggedright
§5.1 (T1-CC column)\strut
\end{minipage} & \begin{minipage}[t]{0.19\columnwidth}\raggedright
§B.1 Table 4 (panel-wide Spearman $\rho$)\strut
\end{minipage} & \begin{minipage}[t]{0.22\columnwidth}\raggedright
$\alpha$ = 0.02; \%-agree = 0.54 (rubric-ambiguity; §A.3)\strut
\end{minipage} & \begin{minipage}[t]{0.15\columnwidth}\raggedright
Resolution-like (Pearson \emph{r} as a proxy): tests whether confidence
orders correctness, not whether the value matches empirical
frequency.\strut
\end{minipage}\tabularnewline
\begin{minipage}[t]{0.06\columnwidth}\raggedright
\textbf{T2-EV} Epistemic Vigilance\strut
\end{minipage} & \begin{minipage}[t]{0.23\columnwidth}\raggedright
§5.1 (T2-EV column)\strut
\end{minipage} & \begin{minipage}[t]{0.19\columnwidth}\raggedright
§B.3 strategy classes; §B.6 DeepSeek pattern\strut
\end{minipage} & \begin{minipage}[t]{0.22\columnwidth}\raggedright
$\alpha$ = 0.13 (below 0.25 bar)\strut
\end{minipage} & \begin{minipage}[t]{0.15\columnwidth}\raggedright
Accuracy on judgements about external claims; no confidence component.
v60/v61 sensitivity table at §5.1.\strut
\end{minipage}\tabularnewline
\begin{minipage}[t]{0.06\columnwidth}\raggedright
\textbf{T3-KB} Knowledge Boundary\strut
\end{minipage} & \begin{minipage}[t]{0.23\columnwidth}\raggedright
§5.1 (T3-KB column, $\ddagger$)\strut
\end{minipage} & \begin{minipage}[t]{0.19\columnwidth}\raggedright
§A.4 v75 bootstrap (6 of 10 items fail to separate high- from
low-scoring models)\strut
\end{minipage} & \begin{minipage}[t]{0.22\columnwidth}\raggedright
$\alpha$ = 0.04; v75 \textbf{exploratory v2 under piloting}\strut
\end{minipage} & \begin{minipage}[t]{0.15\columnwidth}\raggedright
Excluded from all headline composite claims.\strut
\end{minipage}\tabularnewline
\begin{minipage}[t]{0.06\columnwidth}\raggedright
\textbf{T4-CR} Calibration Range\strut
\end{minipage} & \begin{minipage}[t]{0.23\columnwidth}\raggedright
§5.1 (T4-CR column)\strut
\end{minipage} & \begin{minipage}[t]{0.19\columnwidth}\raggedright
§B.2 Flash 12-item table; §A.5 9-config sensitivity sweep\strut
\end{minipage} & \begin{minipage}[t]{0.22\columnwidth}\raggedright
$\alpha$ = 0.66 (only task to clear the $\alpha$ bar)\strut
\end{minipage} & \begin{minipage}[t]{0.15\columnwidth}\raggedright
\%-agree 0.83 \textless{} 0.90 target; $\sigma$\_T / R\_T thresholds embargoed
to OSF.\strut
\end{minipage}\tabularnewline
\begin{minipage}[t]{0.06\columnwidth}\raggedright
\textbf{T5-RCV} Reasoning-Chain Validation\strut
\end{minipage} & \begin{minipage}[t]{0.23\columnwidth}\raggedright
§5.1 (T5-RCV column)\strut
\end{minipage} & \begin{minipage}[t]{0.19\columnwidth}\raggedright
§B.6 DeepSeek 27-pt T2-T5 gap\strut
\end{minipage} & \begin{minipage}[t]{0.22\columnwidth}\raggedright
$\alpha$ = 0.00 (rubric-ambiguity floor)\strut
\end{minipage} & \begin{minipage}[t]{0.15\columnwidth}\raggedright
Accuracy on judgements about reasoning chains; no confidence component.
arXiv v2 per-item audit pending.\strut
\end{minipage}\tabularnewline
\bottomrule
\end{longtable}

\textbf{Where to find the headline.} §5.2 --- the 47-point within-model
Flash dissociation, audited at item level. \textbf{Where to find the
caveats.} §6 (limitations); §A.2 Table 2 (reliability gap analysis).
\textbf{Where to find the replication plan.} §E.3 --- May--June 2026
rubric pilot; Q3 2026 N$\geq$20 model panel + N$\geq$100 stratified humans +
strict-metacognitive sample-variance probes. \textbf{Where to find the
data.} §E.2 --- \texttt{submission\_output/} per-task per-model
\texttt{.run.json}; \texttt{human\_baselines/} 69-rater $\times$ 30-item
matrix.

\begin{center}\rule{0.5\linewidth}{0.5pt}\end{center}

\hypertarget{references}{%
\subsection{References}\label{references}}

Brier, G. W. (1950). Verification of forecasts expressed in terms of
probability. \emph{Monthly Weather Review}, 78(1), 1--3.

Flavell, J. H. (1979). Metacognition and cognitive monitoring: A new
area of cognitive-developmental inquiry. \emph{American Psychologist},
34(10), 906--911.

Guadagnoli, E., \& Velicer, W. F. (1988). Relation of sample size to the
stability of component patterns. \emph{Psychological Bulletin}, 103(2),
265--275.

Guo, C., Pleiss, G., Sun, Y., \& Weinberger, K. Q. (2017). On
calibration of modern neural networks. In \emph{Proceedings of ICML
2017}.

Gwet, K. L. (2014). \emph{Handbook of Inter-Rater Reliability} (4th
ed.). Advanced Analytics, LLC.

Haladyna, T. M. (2004). \emph{Developing and Validating Multiple-Choice
Test Items} (3rd ed.). Lawrence Erlbaum Associates.

Hendrycks, D., Burns, C., Basart, S., Zou, A., Mazeika, M., Song, D., \&
Steinhardt, J. (2021). Measuring massive multitask language
understanding. In \emph{Proceedings of ICLR 2021}.

Kadavath, S., Conerly, T., Askell, A., Henighan, T., Drain, D., et
al.~(2022). Language models (mostly) know what they know.
\emph{arXiv:2207.05221}.

Kane, M. T. (2006). Validation. In R. L. Brennan (Ed.),
\emph{Educational Measurement} (4th ed., pp.~17--64). Praeger.

Krippendorff, K. (2011). Computing Krippendorff's alpha-reliability.
\emph{Annenberg School for Communication Departmental Papers}.

Liang, P., Bommasani, R., Lee, T., Tsipras, D., et al.~(2023). Holistic
evaluation of language models (HELM). \emph{Transactions on Machine
Learning Research}.

Lin, S., Hilton, J., \& Evans, O. (2022a). Teaching models to express
their uncertainty in words. \emph{Transactions on Machine Learning
Research}.

Lin, S., Hilton, J., \& Evans, O. (2022b). TruthfulQA: Measuring how
models mimic human falsehoods. In \emph{Proceedings of ACL 2022}.

Messick, S. (1989). Validity. In R. L. Linn (Ed.), \emph{Educational
Measurement} (3rd ed., pp.~13--103). Macmillan.

Murphy, A. H. (1973). A new vector partition of the probability score.
\emph{Journal of Applied Meteorology}, 12(4), 595--600.

Nelson, T. O., \& Narens, L. (1990). Metamemory: A theoretical framework
and new findings. In G. H. Bower (Ed.), \emph{Psychology of Learning and
Motivation}, 26, 125--173. Academic Press.

Rein, D., Hou, B. L., Stickland, A. C., Petty, J., Pang, R. Y., Dirani,
J., Michael, J., \& Bowman, S. R. (2024). GPQA: A graduate-level
Google-proof Q\&A benchmark. In \emph{First Conference on Language
Modeling (COLM)}.

Srivastava, A., Rastogi, A., Rao, A., Shoeb, A. A. M., et al.~(2023).
Beyond the Imitation Game: Quantifying and extrapolating the
capabilities of language models. \emph{Transactions on Machine Learning
Research}.

Xiong, M., Hu, Z., Lu, X., Li, Y., Fu, J., He, J., \& Hooi, B. (2024).
Can LLMs express their uncertainty? An empirical evaluation of
confidence elicitation in LLMs. In \emph{Proceedings of ICLR 2024}.

\begin{center}\rule{0.5\linewidth}{0.5pt}\end{center}

\emph{End of paper. Compiled length: \textasciitilde12.0 pages under
\texttt{\textbackslash{}documentclass{[}11pt{]}\{article\}}, 1-inch
margins, single column. All technical matter is in §A--§E above, inside
the 10--12-page envelope.}

\end{document}